\documentclass[letterpaper,journal]{IEEEtran}
\usepackage{amsmath,amsfonts}
\usepackage{array}
\usepackage{textcomp}
\usepackage{stfloats}
\usepackage{url}
\usepackage{verbatim}
\usepackage{graphicx}
\usepackage{cite}
\usepackage{amsmath}
\usepackage{cite}
\usepackage{bm}
\usepackage{array}  
\usepackage{threeparttable,multirow,booktabs}
\usepackage[table,xcdraw]{xcolor}
\definecolor{mygray}{gray}{0.85}
\usepackage{bbding}
\usepackage{textcomp}
\usepackage{url,hyperref,bookmark}
\usepackage{pifont}
\usepackage{amssymb}
\usepackage{xspace}
\usepackage{graphicx}
\usepackage{subfigure}
\usepackage{diagbox} 

\usepackage{comment}

\usepackage{array}
\newcommand{\Qian}[1]{{\color{black}{#1}}}
\newcommand{\Huang}[1]{{\color{black}{#1}}}

\begin{document}

\title{
\Qian{Low, High, and Semantic-Level Vision-Language Learning for AI-Generated Image Detection
}}

\author{

\IEEEauthorblockN{Qian Yao, Luming Yang, Jun-Jie Huang, Yongjun Wang, Xinwang Liu, \IEEEmembership{Senior Member, IEEE}, and Meng Wang, \IEEEmembership{Fellow, IEEE}
}

\thanks{Qian Yao, Jun-Jie Huang, Yongjun Wang, and Xinwang Liu are with the College of Computer Science and Technology, National University of Defense Technology, Changsha 410073, China (e-mail:
yaoqian21@nudt.edu.cn; jjhuang@nudt.edu.cn; wwyyjj1971@126.com; xinwangliu@nudt.edu.cn.) Luming Yang is with the Academy of Military Science, Beijing 100091, China (e-mail: yang\_lm108@nudt.edu.cn.) Meng Wang is with the Hefei University of Technology, Hefei 230009, China (e-mail: wangmeng@hfut.edu.cn) (\textit{Corresponding authors: Jun-Jie Huang, and Yongjun Wang})}
}

\markboth{}%
{\Huang{Yao} \MakeLowercase{\textit{et al.}}: \Huang{LHSDet: AI-Generated Image Detection via Low- and High-Level Visual and Semantic Textual Feature Fusion
}}

\maketitle
\begin{abstract}
Driven by advances in diffusion models and autoregressive models, the fidelity and resolution of AI-generated images now rival those of real images. However, existing AI-generated image detection methods often downsample the images, inevitably overlooking critical low-level texture details in high-resolution AI-generated images, \Huang{thereby} limiting their detection performance.
In addition, the \Huang{continuous} emergence of unknown generative models makes large-scale pre-training datasets inaccessible. To address these challenges, we propose a novel AI-generated image detector, termed \Huang{LHSDet (\textbf{L}ow-, \textbf{H}igh-, and \textbf{S}emantic-level \textbf{Det}ection)}. Specifically, we formulate the AI-generated image detection task as a Visual Question Answering (VQA) problem, leveraging a fine-tuned vision-language framework to fully exploit the complementary information between visual and textual modalities. 
Recognizing that the default visual encoder of existing vision-language models is not tailored for AI-generated image detection, we redesign a visual encoder to better capture both the low-level and high-level artifacts inherent in AI-generated images.
Furthermore, we incorporate a semantic-level textual branch to enable multi-modal feature fusion and detection. 
Consequently, LHSDet employs a triple-branch architecture to extract complementary multi-modal features: a low-level visual branch that aggregates non-overlapping patches for local texture cues, a high-level visual branch based on SigLIP2 for global perception feature extraction, and a semantic-level textual branch that generates captions using BLIP-2.
Extensive experimental results demonstrate that LHSDet achieves state-of-the-art (SOTA) detection accuracy and robust performance across diverse generative models, including both diffusion and autoregressive models, as well as on the WildRF benchmark. \Huang{Notably, on the high-resolution benchmark, LHSDet outperforms the second-best baseline by 9.85\% in average accuracy and improves the worst-case accuracy across ten unseen generators from 70.33\% to 85.33\%, validating its effectiveness and superiority in addressing the aforementioned challenges.}

\end{abstract}


\section{Introduction}

\IEEEPARstart{T}{he} proliferation of AI-generated content is profoundly impacting a wide range of applications. \Qian{As AI-generated videos on social networks often exhibit frame-level forgeries,} the underlying image generation mechanisms have evolved significantly, progressing from the early dominant methods like
Generative Adversarial Networks (GANs)~\cite{gan} and Variational Autoencoders (VAEs)~\cite{vae} to the more recent paradigm established by Denoising Diffusion Probabilistic Models (DDPM)~\cite{ddpm}.
This breakthrough has spurred the development of advanced methods~\cite{adm,vqdiffusion,pndm,imagen} such as Latent Diffusion Models (LDM)~\cite{sd}, Stable Diffusion~\cite{sd}, DALL·E2~\cite{dalle2}, \textit{et al}. And recent diffusion models now enable the synthesis of high-resolution images
~\cite{pixart,sana,ultrapixel}.

These increasingly realistic AI-generated images have rendered them nearly indistinguishable from natural photographs. However, the malicious exploitation of such synthetic images can \Huang{pose} serious societal threats, including the potential interference in political elections, manipulation of public opinion, violation of individual privacy, and plagiarism of intellectual property. 
The emergence of unknown AI image generation tools further complicates the detection landscape.
Therefore, the development of effective detection methods that can generalize to new generators with limited training data has become an urgent research imperative.

\Huang{AI-generated image detection methods exploit the characteristic differences between AI-generated and natural images, which exist at different levels.}
Existing approaches for AI-generated image detection can be broadly classified into three categories: Convolutional Neural Networks (CNN)-based detectors~\cite{upsampling,lgrad,dualnet,f3net,corv}, Reconstruction Error (RE)-based detectors~\cite{dire,aero,drct,fakeinversion,ronan,lare2,zerofake} and Vision-language model (VLM)-based detectors.
(i) The CNN-based detectors~\cite{upsampling,lgrad,dualnet,f3net,corv} typically examine low-level cues or artifacts presented in AI-generated images. Though effective, they tend to overfit to specific artifacts seen during training, leading to poor generalization and significant performance degradation in few-shot scenarios. Moreover, their robustness is often limited and the efficacy could drop significantly when the testing images undergo image post-processing operations. 
(ii) The RE-based detectors~\cite{dire,aero,drct,fakeinversion,ronan,lare2,zerofake} \Huang{leverage the observation that an AI-generated image can be reconstructed by the corresponding generative model with a smaller error than a real image.}
Detection can therefore be performed by comparing an input image to its reconstruction, under the assumption that generated images will yield a smaller reconstruction error. 
However, this approach is computationally intensive
due to the iterative nature of diffusion models. Furthermore, the detection efficacy is 
heavily \Huang{dependent} on the selected reconstruction model, resulting in degraded cross-model performance.
(iii) 
The VLM-based detectors typically leverage large pretrained models like CLIP~\cite{clip,univfd,defake,fatformer,c2pclip,aide} and BLIP~\Huang{\cite{blip,antifake}} for their strong semantic understanding capability. However, the input images are usually downsampled to a lower resolution for processing~\cite{univfd,fatformer},
overlooking the fine-grained texture and edge information crucial for high-resolution image detection. 
Moreover, since the visual encoders in these models are designed for general-purpose vision tasks, they are not inherently optimized for the specific artifacts of generated image, limiting their detection effectiveness.

In this paper, we extract complementary \textbf{L}ow-, \textbf{H}igh- and \textbf{S}emantic-level cues for high-resolution AI-generated image \textbf{det}ection, termed \textbf{LHSDet}.
The model is built upon a triple-branch architecture.
The first branch is designed to preserve fine-grained texture cues
via non-overlapping patch aggregation
 avoiding the information loss. 
The second branch is dedicated to capturing high-level visual cues essential for generalization in data-scarce scenarios.
This branch \Huang{utilizes} a vision-language model \Huang{that} is effectively fine-tuned for \Huang{the} AI-generated image detection task using Low-Rank Adaptation (LoRA).
\Huang{The outputs} of the first two branches are \Huang{fused} through a cross-attention mechanism, allowing effective integration of low-level and high-level cues.
The third branch introduces a semantic dimension by generating a textual caption of the input image. 
It is built upon a Visual Question Answering (VQA) framework, which takes the fused visual and textual features as input and generates an answer as output.

\Huang{The architecture of LHSDet follows a one-to-one mapping between the observed challenges and the three branches. (i) Since downsampling irreversibly removes high-frequency texture artifacts (Fig.~\ref{fig_inconsistency} (a) and (b)), the low-level visual branch processes non-overlapping patches at the native resolution without any downsampling. (ii) Since few-shot training causes CNN-based detectors to overfit to seen artifacts, the high-level visual branch adapts a pre-trained SigLIP2 encoder via LoRA to provide generalizable global perception. (iii) Since purely visual cues cannot capture semantic ambiguity and physical implausibility (Fig.~\ref{fig_inconsistency} (c) and (d)), the semantic-level textual branch translates the image into a caption so that the LLM can reason over complementary visual and textual evidence.}

The main contributions are summarized as follows:

\begin{itemize}
\item \Qian{We propose LHSDet, which redesigns the visual encoder within the VLM framework, replacing downsampling with native-resolution patch aggregation in the low-level branch and fusing this branch with the high-level \Huang{perception} branch, and further introduces a textual semantic branch as auxiliary evidence for detection.}

\item LHSDet
leverages multi-modal feature fusion of visual and textual inputs to attain \Huang{robust detection with parameter-efficient fine-tuning in which only 0.23\% of the parameters are trainable}.

\item 
\Huang{Extensive experimental results demonstrate that LHSDet outperforms SOTA baselines by 9.85\% in average accuracy on the high-resolution benchmark and by 6.56\%/0.76\% on the pixel/semantic subsets of AIGI-Now, and improves the worst-case accuracy across ten unseen generators from 70.33\% to 85.33\%, while also demonstrating strong robustness.}

\end{itemize} 

The rest of the paper is organized as follows: 
Section~\ref{sec:rw} reviews the related works in AI-generated image detection.  Section~\ref{sec:mt} describes the design details of the proposed LHSDet. Section~\ref{sec:exp} presents the implementation details and discusses the experimental results. Finally, Section~\ref{sec:con} concludes the paper.





\section{Related work}
\label{sec:rw}
In this section, we introduce the approaches of AI-generated image detection, which can be categorized into three main types: the CNN-based, the RE-based, and the VLM-based methods.

\subsection{The CNN-based Detection Methods}
Traditional methods primarily leverage spatial features and frequency artifacts inherent in generated images, such as color~\cite{upsampling}, gradients~\cite{lgrad}, noise patterns~\cite{dualnet}, and frequency artifacts~\Huang{\cite{f3net,corv}}.
For instance, NPR~\cite{upsampling} detects upsampling artifacts in generative models by examining the neighboring pixel relationships. \Huang{LGrad~\cite{lgrad} employs a pretrained CNN to transform images into gradient maps that reveal generalized artifacts, and then feeds these maps into a classifier.} Xi \textit{et al.}~\cite{dualnet} propose a dual-stream network designed to extract high-frequency texture information via a residual stream, while using a content stream to capture forged traces in low-frequency components.
Frequency domain analysis has proven particularly effective, as generative models often leave distinctive spectral artifacts. 
Frequency in Face Forgery Network (F3Net)~\cite{f3net} utilizes \Huang{frequency-aware} forgery clues. Similarly, CNNDet~\cite{cnnd} utilizes a pre-trained ResNet50 as a binary classifier to examine whether there are common patterns in the Fourier domain of GAN-generated images. However, as noted by Corvi \textit{et al.}~\cite{corv}, frequency artifacts in recent diffusion models are less noticeable compared to those in GANs, leading to performance degradation for methods overly reliant on spectral signatures. 
A significant limitation of many spatial and frequency-based CNN classifiers is their vulnerability to image perturbations like Gaussian blur and their limited generalization ability to unseen generative models or in few-shot scenarios.
To address these limitations, \Huang{an} alternative approach~\cite{shadows} leverages projective geometry by detecting physical inconsistencies such as anomalous shadow\Huang{s} or vanishing point\Huang{s} in AI-generated images.
While offering a novel, physically-grounded perspective, this method lacks generalizability as not all images contain such explicit geometric elements.

\subsection{The RE-based Detection Methods}
A distinct and increasingly prominent approach for detecting AI-generated images, particularly those from diffusion models, leverages the reconstruction error inherent to the generative process itself~\cite{dire,drct,fakeinversion,ronan,aero,lare2,zerofake}.
The core hypothesis is that images synthesized by a given diffusion model are more easily reconstructed by that same model than natural images.
Diffusion Reconstruction Error (DIRE)~\cite{dire} employs a Denoising Diffusion Implicit Model (DDIM)~\cite{ddim} to measure the error between an image and its reconstruction. 
While effective, DIRE suffers from inefficient sampling and requires lengthy training duration.
Chen \textit{et al.}~\cite{drct} propose a diffusion reconstruction contrastive training method for the universal detection of diffusion-generated images. FakeInversion~\cite{fakeinversion} enhances detection by leveraging a richer representation which composing the original image, the approximate noise map obtained via text-conditioned DDIM inversion with Stable Diffusion, and its corresponding reconstruction from denoising. These are then fed into a ResNet50 classifier. RONAN~\cite{ronan} develops a model-agnostic and alteration-free framework for attributing image origins via reverse engineering of generative models. 
AEROBLADE~\cite{aero} adopts a training-free strategy for latent diffusion images using autoencoder reconstruction errors combined with a Learned Perceptual Image Patch Similarity (LPIPS) metric computed on image patches.

Though effective, reconstruction-based methods face two primary limitations: (i) 
the necessity of running a diffusion model renders the detection computationally expensive and time-consuming, and 
(ii) their performance is intrinsically dependent on the model used for reconstruction, which may not generalize well to images produced by unknown generative models.

\subsection{The VLM-Based Detection Methods}
Current approaches to visual-language model detection predominantly employ CLIP~\cite{clip,univfd,defake,fatformer,c2pclip,aide} and BLIP~\cite{blip,antifake} as foundational models. 
The core paradigm exploits the semantic gap between real and generated content within a shared multimodal embedding space. 
This can be broadly categorized into two strategies: methods that measure image-text similarity and those that reframe detection as a Visual Question Answering (VQA) task.
UnivFD~\cite{univfd} leverages a pre-trained CLIP ViT network to map both real and generated images into their corresponding feature representations, using cosine distance to compare a query image against a database of real and fake feature templates for classification. 
Tang \textit{et al.}~\cite{cail} proposes an adapter-based domain incremental learning framework utilizing a pre-trained ViT for AI-generated image detection. Defake~\cite{defake} leverages CLIP’s image and text encoders to project images and prompts into a shared embedding space, detecting generated image by measuring the distance between prompt and image embeddings. FatFormer~\cite{fatformer} proposes a forgery-aware adapter with language-guided alignment, which first extracts low-level forgery traces, then aligns the image features to text-prompt embeddings. C2P-CLIP~\cite{c2pclip} incorporates a category common prompt into the text encoder to introduce category-related concepts into the image encoder, thereby improving detection performance. AIDE~\cite{aide} fuses DCT-based low-level cues with CLIP-driven high-level visual embeddings via a Mixture-of-Experts (MoE) architecture for AI-generated image detection. Keita \textit{et al.} ~\cite{harness} investigate the effectiveness of advanced vision-language models such as BLIP-2 and ViT-GPT2 for the detection of generated images. AntifakePrompt~\cite{antifake} reformulates generated-image detection as a VQA task, implemented via an InstructBLIP~\cite{instructblip} vision–language model (VLM) and Vicuna-7B large-language model (LLM). A learnable pseudo-token is inserted into the input prompt, with only its embedding being optimized while keeping all pre-trained parameters frozen. 

While promising, VLM-based detectors face significant challenges. The default visual encoders of models like CLIP are not inherently designed to capture subtle, low-level forgery artifacts critical for detection. Furthermore, the common practice of downsampling high-resolution input images to meet the model's fixed input size often leads to the loss of fine-grained texture details that are essential for robust forensic analysis.

\section{Methodology}
\label{sec:mt}

\begin{figure*}
    \centering
    \includegraphics[width=0.98\linewidth]{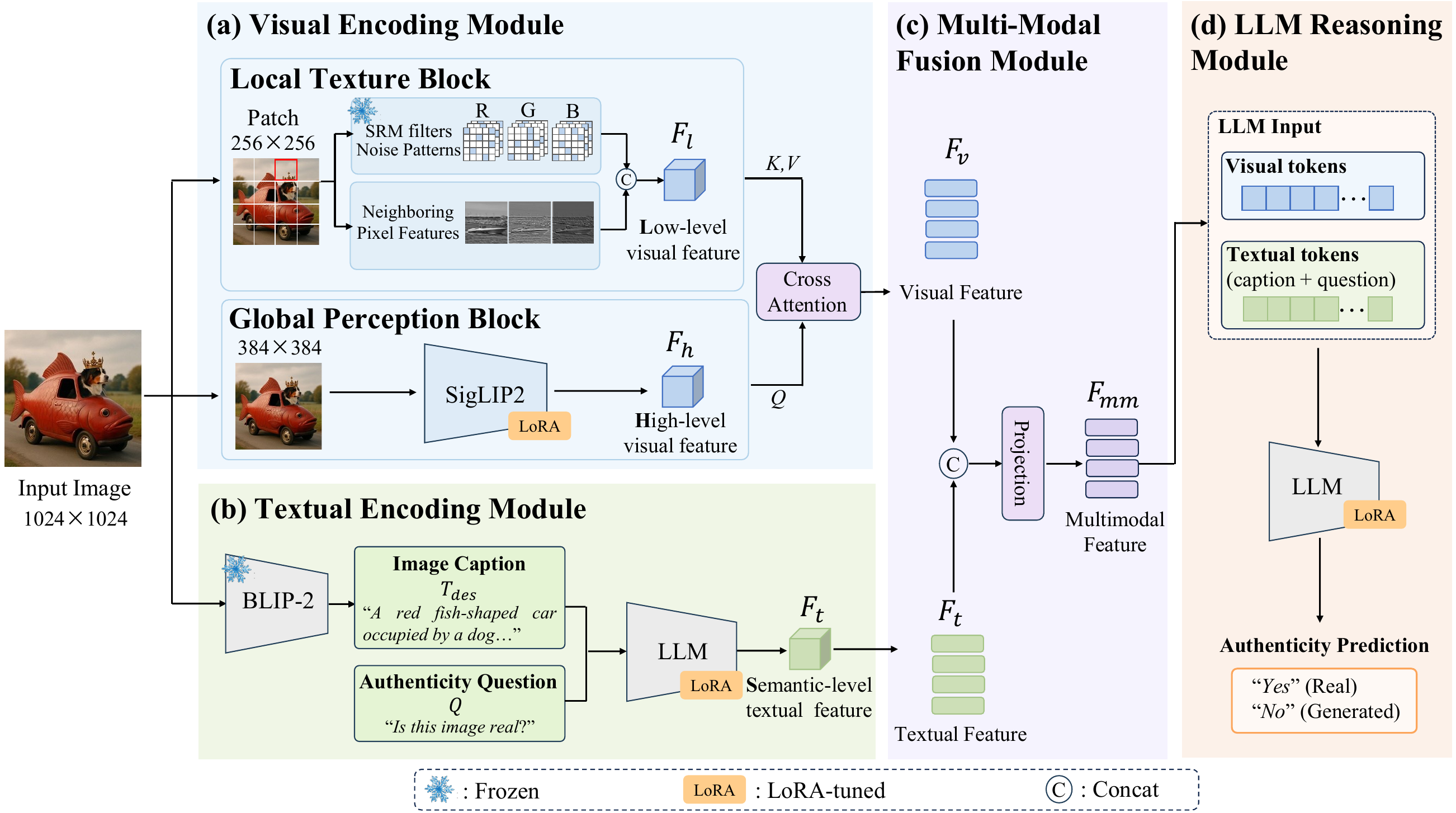}
    \caption{
    Overview of the proposed LHSDet, which comprises a Visual Encoding Module (VEM), a Textual Encoding Module (TEM), a Multi-Modal Feature Fusion Module (MFM), and an LLM Reasoning Module (LRM). (a) VEM: the \textit{low-level branch} divides the input image into non-overlapping patches to capture texture cues, SRM noise patterns and neighboring-pixel inconsistencies, without full-image downsampling, while the \textit{high-level branch} encodes a downsampled global view with the LoRA-adapted SigLIP2 encoder; the two branches are fused via global-guided cross-attention. (b) TEM as a \textit{Semantic-level branch:}  BLIP-2 generates a caption for the input image, which is concatenated with the question as explicit semantic evidence. (c) MFM: the fused visual feature and the textual feature are packed into aligned visual and textual tokens. (d) LRM: the LLM produces an answer ("\texttt{Yes}"/"\texttt{No}") to the question "\texttt{Is this image real?}". 
    }
    \label{fig_LHSDet}
\end{figure*}


\Huang{The 
AI-generated image detection task confronts two main challenges. First, 
AI-generated images often have to be downsampled for compatibility with standard detection pipelines, and this preprocessing step irretrievably loses the subtle texture artifacts. Second, the emergence of unknown generative models imposes a few-shot constraint, and the scarcity of annotated data prevents existing detectors from generalizing to diverse, unseen synthetic content.}

To mitigate these challenges, we propose LHSDet, a novel triple branch detector that extracts \textbf{L}ow-level and \textbf{H}igh-level vision features along with \textbf{S}emantic-level textual features for high-resolution generated image \textbf{det}ection. Overall, we reformulate AI-generated image detection as a VQA task to fully harness the \Huang{strong few-shot generalization capability of VLMs}. 
Concurrently, considering that the default visual encoder of existing VLM is not tailored for generated image detection, we redesign a visual encoder that incorporates both low-level local texture and high-level global feature extraction rather than directly employing the default visual encoder. 
Furthermore, in the semantic-level, we generate a caption for the input image which is \Huang{concatenated} with the question to form the semantic textual feature.
Finally, \Huang{an} LLM yields the detection results by effectively fusing the low-level, high-level and semantic-level representations. \Huang{Formally, the detection task is reformulated as answering a preset question about the image: given an input image $\mathbf{X}$ and a question $q$ (\textit{e.g.}, ``\texttt{Is this image real?}''), a vision--language model $\mathcal{M}$ produces an answer $y \in \{\texttt{Yes}, \texttt{No}\}$, \textit{i.e.}, $\mathcal{M}(\mathbf{X}, q) \rightarrow y$, which corresponds to the binary decision of AI-generated image detection.}

\Huang{The design of LHSDet follows three principles, each addressing one of the challenges above. (P1) \textit{Asymmetric resolution allocation}: texture evidence is resolution-fragile, whereas semantic evidence is resolution-robust; the low-level branch therefore operates on non-overlapping patches at the native resolution without any downsampling, while the high-level branch exploits a downsampled global view for efficiency. (P2) \textit{Frozen priors with learnable lightweight adaptation}: handcrafted artifact priors (SRM) and foundation-model components (SigLIP2, BLIP-2, and the LLM backbone) are kept frozen to resist few-shot overfitting and catastrophic forgetting; only the convolutional refinement layers, the LoRA adapters, and the fusion modules are learned, accounting for merely 0.23\% of all parameters. (P3) \textit{Evidence-oriented multimodality}: instead of using text as auxiliary supervision, the input image is translated into a caption that serves as explicit semantic evidence, and the LLM weighs this evidence against the visual features through a posterior-calibrated Yes/No answer (Sec.~\ref{sec:lrm}).}

\Huang{Among these components, the SRM filters, the SigLIP2 encoder, BLIP-2, and the LoRA technique are inherited from prior work, whereas the new designs of LHSDet are: (i) the native-resolution patch aggregation embedded into the VLM visual front-end, (ii) the global-guided cross-attention that fuses the low- and high-level visual branches, (iii) the caption-as-evidence question construction, and (iv) the posterior-calibrated answer scoring. Table~\ref{tab_property} positions LHSDet against the closest detectors along these dimensions.}

\Huang{
\begin{table}[t]
\caption{\Huang{
Comparison of LHSDet with the closest AI-generated image detectors. ``Nat.-Res'' denotes processing the input at its native resolution without downsampling; ``Low-Lvl'' denotes an explicit low-level artifact branch; ``Text.'' denotes textual semantic evidence about the image content; ``Param.'' denotes training-free or parameter-efficient adaptation.
}}
\begin{center}
\Huang{\small}
\setlength{\tabcolsep}{3.5pt}
\begin{tabular}{l|ccccc}
\toprule
\Huang{\textbf{Method}} & \Huang{\textbf{Nat.-Res}} & \Huang{\textbf{Low-Lvl}} & \Huang{\textbf{Text.}} & \Huang{\textbf{VQA}} & \Huang{\textbf{Param.}} \\
\midrule
\Huang{UnivFD~\cite{univfd}} & \ding{55} & \ding{55} & \ding{55} & \ding{55} & \ding{51} \\
\Huang{AEROBLADE~\cite{aero}} & \ding{51} & \ding{55} & \ding{55} & \ding{55} & \ding{51} \\
\Huang{AIDE~\cite{aide}} & \ding{55} & \ding{51} & \ding{55} & \ding{55} & \ding{55} \\
\Huang{AntifakePrompt~\cite{antifake}} & \ding{55} & \ding{55} & \ding{55} & \ding{51} & \ding{51} \\
\midrule
\Huang{LHSDet (ours)} & \ding{51} & \ding{51} & \ding{51} & \ding{51} & \ding{51} \\
\bottomrule
\end{tabular}
\end{center}
\label{tab_property}
\vspace{-0.3cm}
\end{table}
}




Fig.~\ref{fig_LHSDet} presents the overview of LHSDet. It comprises four main modules, \textit{i.e.}, a Visual Encoding  Module (VEM), a Textual Encoding Module (TEM), a Multi-Modal Feature Fusion Module (MFM) and an LLM Reasoning Module (LRM). 
The VEM extracts complementary visual features 
from two parallel processing branches, a local texture block and a global perception block. 
The local texture block processes non-overlapping image patches to capture low-level cues including neighboring pixel inconsistency and noise patterns.
The global perception block 
uses 
SigLIP2's~\cite{siglip2} visual encoder to capture high-level visual features\Huang{, such as} object-level inconsistencies.
These two visual features are fused by cross-attention mechanism. 
The TEM constructs semantic-level textual features using BLIP-2 to generate a detailed caption for the input image.
This caption combines with the question to form a semantic-level textual feature. 
The MFM generates a unified multi-modal representation by integrating the VEM's visual feature and TEM's textual feature
through a structured pipeline. It enables simultaneous attention to both modalities while maintaining dimensional consistency. 
The unified multi-modal feature from MFM is formatted into aligned visual/textual tokens, and the LRM processes these tokens to reason about the image’s authenticity and output a binary answer (``\texttt{Yes}"/``\texttt{No}").
\Huang{In short, the four modules implement the three complementary branches: the VEM hosts the low-level and high-level visual branches, the TEM hosts the semantic-level textual branch, and the MFM and LRM fuse the three branches and produce the final answer.}

\subsection{Visual Encoding Module: Low- and High-Level Visual Branches}
\Qian{Recent generative models~\cite{pixart,sana,ultrapixel} are capable of generating high-resolution synthetic images.}
However, prevailing image encoders~\cite{univfd,fatformer,antifake} still down-sample inputs to \Huang{a} resolution of $256 \times 256$ for pre-processing, discarding the fine-grained details and sharp edges of the AI-generated images. 
Meanwhile, 
the inconsistency between generated and real images \Huang{goes} beyond low-level texture features, such as color and noise, to critically involve high-level visual characteristics, most notably semantic ambiguity and violations of physical logic.
To address this gap, 
we propose the Visual Encoding Module (VEM) with a dual-branch architecture, enabling \Huang{the joint capture of}
AI-generated artifacts by extracting features at both low-level and high-level.
The local texture features and global perception features are combined to obtain the fused visual features. The texture and perception streams complement each other, jointly enabling comprehensive detection of synthetic images. \Huang{In contrast to directly employing the default VLM visual encoder, the VEM constitutes a detection-oriented visual front-end: the low-level branch preserves resolution-fragile texture evidence at the native resolution, whereas the high-level branch extracts resolution-robust semantics from a downsampled global view (P1).}

\subsubsection{Local Texture Block: Native-Resolution Patch Aggregation}
\begin{figure*}
    \centering
    \subfigure[]{
        \includegraphics[width=0.22\linewidth]{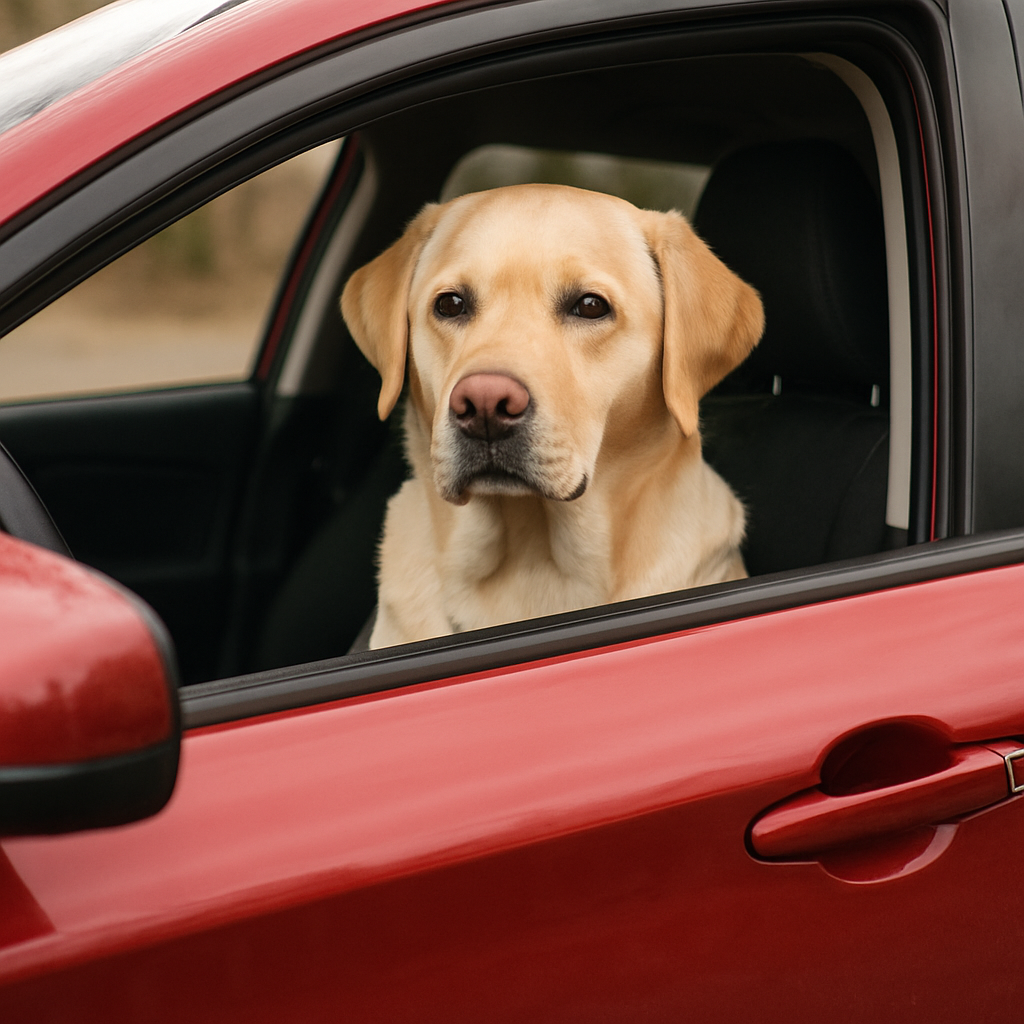}
        \label{fig:sub1}
    }
    \subfigure[]{
        \includegraphics[width=0.22\linewidth]{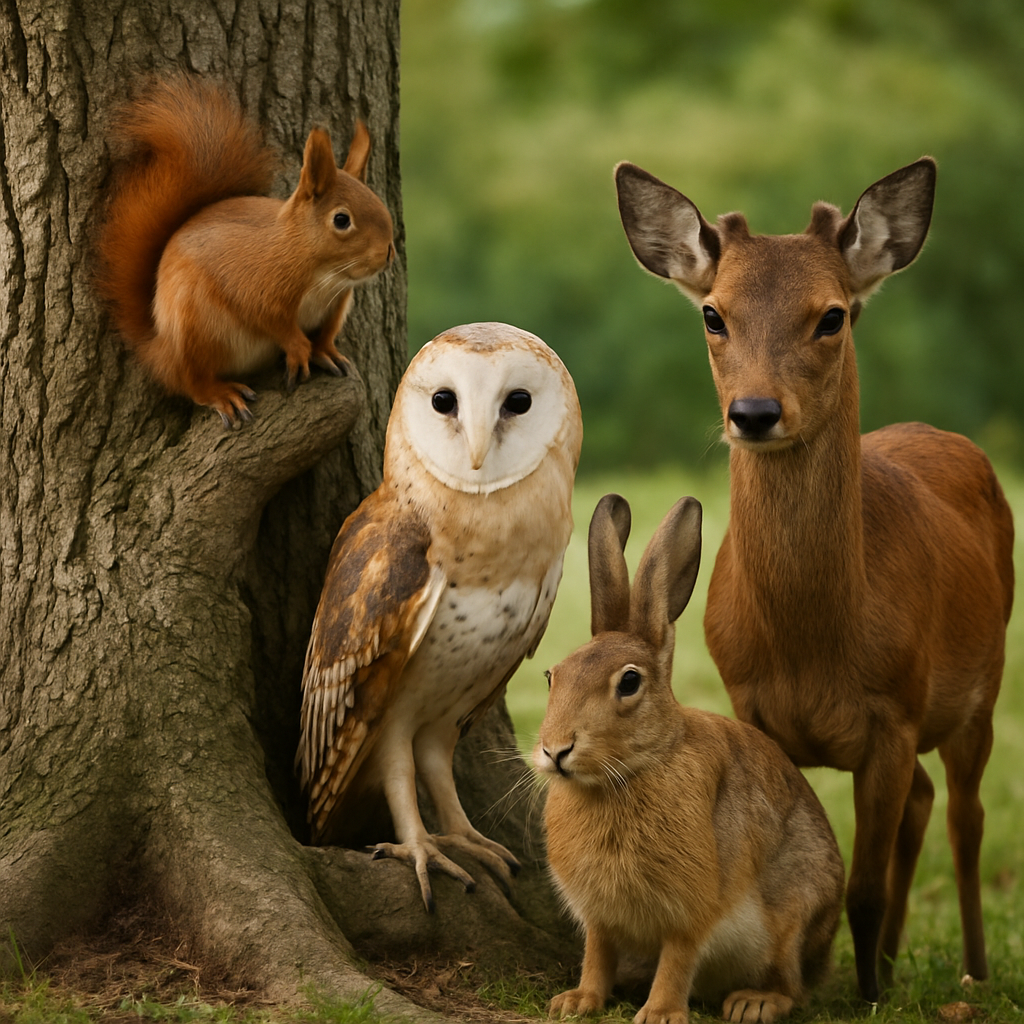}
        \label{fig:sub2}
    }
     \subfigure[]{
        \includegraphics[width=0.22\linewidth]{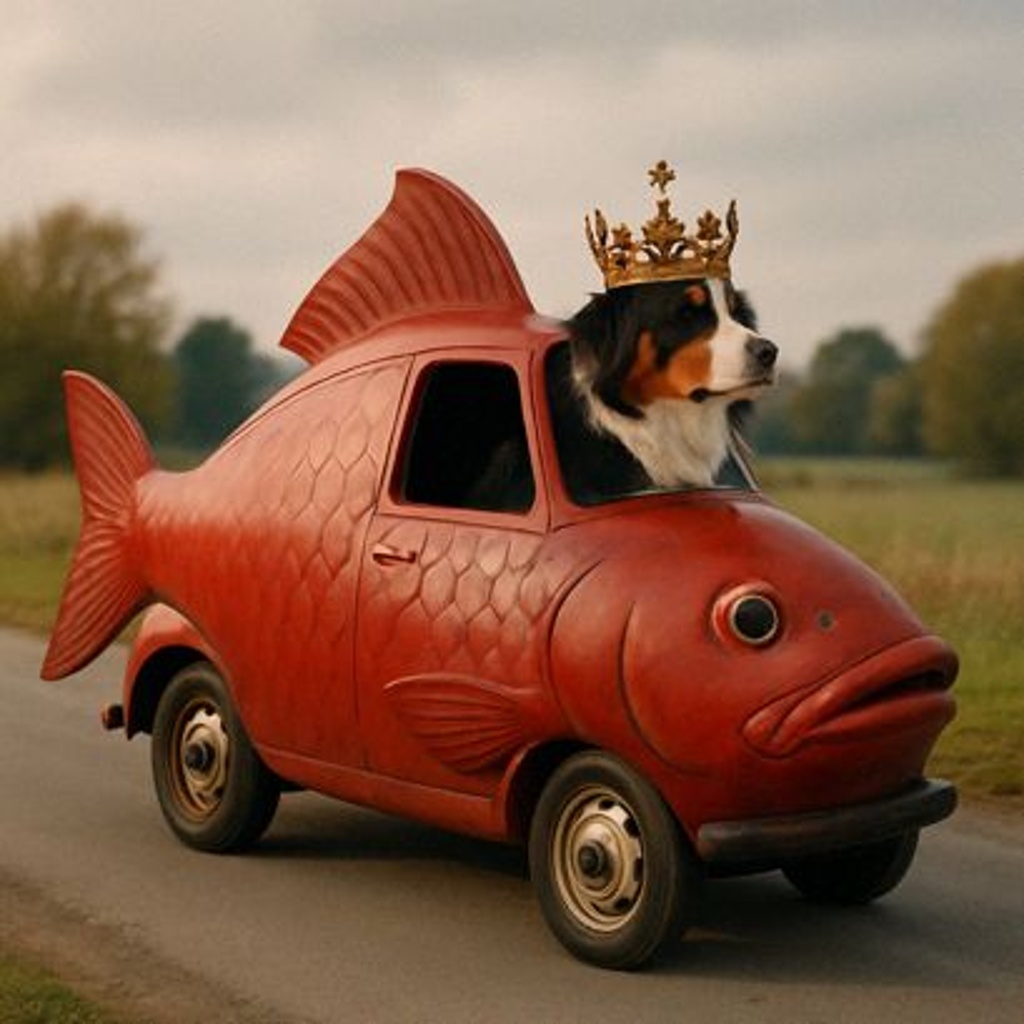}
        \label{fig:sub3}
    }
    \subfigure[]{
        \includegraphics[width=0.22\linewidth]{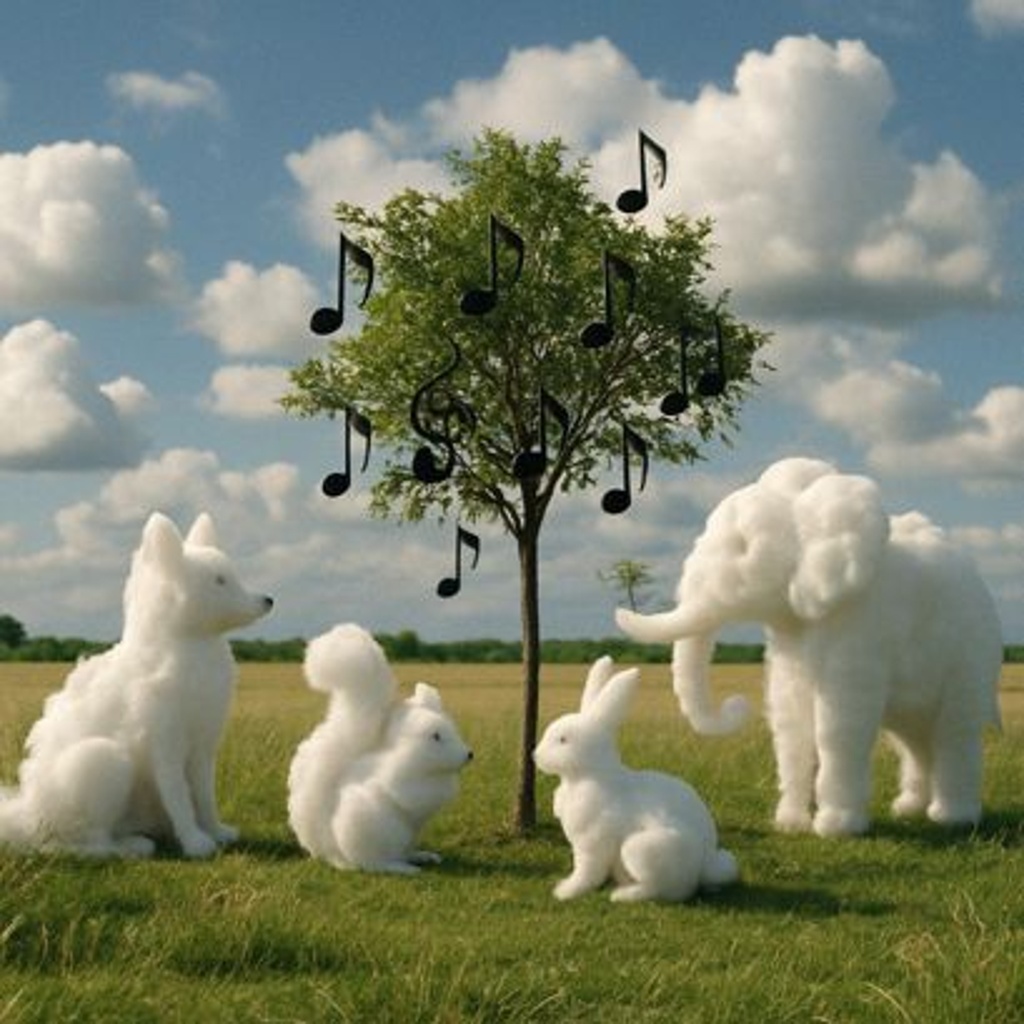}
        \label{fig:sub4}
    }
    
    
    \caption{
    \Qian{Representative examples generated by GPT-4o, illustrating two types of visual inconsistencies in AI-generated images. (a) and (b): low-level inconsistencies in texture and edges, which can only be reliably detected with fine-grained texture features. (c) and (d): high-level semantic inconsistencies, which are more effectively captured by the vision-language model's global visual features.}
    }
    \label{fig_inconsistency}
\end{figure*}

\Huang{Following (P1), this branch processes the input at its native resolution, since downsampling would irreversibly remove the high-frequency artifacts that distinguish generated images.} This block captures the low-level visual features, such as noise patterns and neighboring pixel relationships. \Huang{As shown in Fig.~\ref{fig_inconsistency} (a) and (b), unnatural smoothness and abrupt changes between neighboring pixels can be observed in the generated images.}

Let us denote an input image as $\mathbf{X} \in \mathbb{R}^{C \times H \times W}$ with $C$ channels and of spatial resolution $H \times W$. The local texture block divides $\mathbf{X}$ into non-overlapping patches $\mathbf{X}_p$ to preserve the texture details while disrupting global semantics~\cite{breaking}, each of resolution $h \times w$. \Huang{Within each patch, the noise patterns are captured via the Spatial Rich Model (SRM)~\cite{srm}, and the neighboring pixel relationships are captured through convolution.} We utilize SRM, a set of high-pass filters tailored for noise-pattern extraction, to obtain the noise representation $\mathbf{F}_\text{noise}$.  Keeping it frozen for noise-pattern extraction ensures high efficiency and mitigates overfitting.

It has been empirically confirmed that up-sampling stages in image-generation pipelines tend to leave distinct artifacts, particularly within the high-frequency components~\cite{corv}. Consequently, the neighboring pixels in generated images may exhibit inconsistencies~\cite{upsampling}. Given its effectiveness in extracting neighboring pixel relationships with low-complexity for superior detection, we follow the prior work~\cite{dualnet} to extract such features. A convolution is performed on the input image, and its output is concatenated with the image patch $\mathbf{X}_p$, yielding a feature map $\mathbf{F}_1$.
A further convolution on $\mathbf{F}_1$ produces the next feature map. 
The two feature maps are concatenated before undergoing a convolution to ultimately generate the final neighboring pixel feature map. 

\begin{equation}
\begin{aligned}
\mathbf{F}_{1} &= [\mathbf{X}_p;\text{Conv}_1(\mathbf{X}_p)], \\
\mathbf{F}_{\text{pixel}} &= \Huang{\text{Conv}_3}\big([\mathbf{F}_{1};\text{Conv}_2(\mathbf{F}_{1})]\big).
\end{aligned}
\end{equation}
\Huang{where $\text{Conv}_1$, $\text{Conv}_2$, and $\text{Conv}_3$ are learnable convolutional layers with distinct parameters}

The two low-level visual features are merged via concatenation along the channel dimension.
Expanding local-texture channels increases model capacity and thus the risk of overfitting under limited data, whereas convolutional fusion adds no parameters. Therefore, we aggregate these low-level visual patches through convolutional refinement rather than channel concatenation. 
\begin{equation}
    {\mathbf{F}}_{l} = \text{Conv}_{p}\big([\mathbf{F}_\text{noise};\mathbf{F}_\text{pixel}]\big).
\end{equation}

\subsubsection{{Global Perception Block: Semantics from a Downsampled Global View}} \Huang{Following (P1) and (P2), this branch captures generalizable high-level semantics with a pre-trained encoder adapted by LoRA, which resists overfitting under the few-shot constraint.} This block aims to capture complementary high-level visual features, including semantic inconsistency and physical implausibility which \Huang{can}not \Huang{be effectively} captured by the local texture block. For instance, as shown in  Fig.~\ref{fig_inconsistency} (c), 
\Qian{a dog is seated on a fish-shaped vehicle.} 
And in Fig.~\ref{fig_inconsistency} (d),
\Qian{the clouds form images of various small animals.}

We extract high-level features with SigLIP2~\cite{siglip2}’s vision encoder with Low-Rank Adaptation (LoRA) adaptation.
The prevailing CLIP-based methods extract coarse global features lacking spatial detail. In contrast, SigLIP2 preserves local semantics through self-supervised patch-level decoding. Meanwhile, image downsampling still maintains high-level feature quality while substantially reducing computational overhead. By default, SigLIP2 processes $1024\times1024$ images at $384\times384$ resolution. 
The SigLIP2's encoder is able to extract high-level visual feature and preserve fine-grained spatial representations. We further fine-tuned the encoder with LoRA to better adapt it for extracting high-level visual features from AI-generated images:
\begin{equation}
    \mathbf{F}_{h} = f_{\text{SigLIP2}}(\mathbf{X}_{384\times384}) \in \mathbb{R}^{D_g},
\end{equation}
where $D_g$ is the global feature dimension, \Huang{$\mathbf{F}_{h}$ is a single global descriptor obtained by pooling the patch tokens of the SigLIP2 encoder.}

\subsubsection{Global-Guided Cross-Attention Fusion}
The high-level visual feature and the low-level visual feature are then fused into a unified visual representation. To enhance mutual interaction between the \Huang{two branches}, they are fused via cross-attention \Huang{along the token dimension}.
We identify the high-level feature from the global perception block as the more essential information which is complemented by the detailed feature from the local texture block. \Huang{The attention direction is deliberate: the global descriptor acts as a single, semantically anchored query that retrieves relevant texture evidence from all local patch tokens, whereas the reverse direction would have to aggregate many noisy local queries; a gated-fusion alternative is compared in Table~\ref{tab_ablation}.}
Specifically, we utilize the global feature to form query, and use the local feature to form key and value:
\begin{equation}
    \begin{aligned}
        \mathbf{Q} &= \mathbf{W}_q  \mathbf{F}_{h} \in \mathbb{R}^{H_d} \\
        \mathbf{K} &= \mathbf{W}_k  {\mathbf{F}}_{l} \in \mathbb{R}^{(h'w')  \times H_d } \\
        \mathbf{V} &= \mathbf{W}_v  {\mathbf{F}}_{l} \in \mathbb{R}^{(h'w')  \times H_d },
    \end{aligned}
\end{equation}
where $H_d$ is \Huang{the} hidden dimension, $\mathbf{W}_q$, $\mathbf{W}_k$, $\mathbf{W}_v$ are projection matrices, \Huang{and} $h'w'$ is
the \Huang{number of local patch tokens}.

The local and global feature fusion is performed as follows:
\begin{equation}
\mathbf{F}_{\text{out}} = \text{Softmax}\left(\frac{\mathbf{Q} \mathbf{K}^\text{T}}{\sqrt{d_k}}\right) \mathbf{V} \in \mathbb{R}^{H_d},
\end{equation}
where \Huang{$d_k=H_d$ and} the scaling factor $\sqrt{d_{\text{k}}}$ stabilizes gradient flow.

Then the fused feature is transformed and integrated with the original global feature:
\begin{equation}
\mathbf{F}_{{v}} = \text{LayerNorm}(\mathbf{W}_{\text{out}}  \mathbf{F}_{\text{out}} + \mathbf{F}_{h}) \in \mathbb{R}^{D_g},
\end{equation}
where $\mathbf{W}_{\text{out}}: \mathbb{R}^{H_d} \rightarrow \mathbb{R}^{D_g}$ projects the fused features back to the global feature space.

This architecture preserves original global information while integrating discriminative details extracted from local features. The visual encoder of LHSDet implements a global-guided local attention mechanism. The connection effectively creates a bidirectional information exchange between global context and spatial details. This symbiotic interaction enables the visual encoder to concurrently maintain high-level visual representations while augmenting its perceptual capacity for low-level features such as texture details.

\subsection{Textual Encoding Module: Caption as Semantic Evidence}
Relying solely on visual features for detection has inherent limitations\Huang{, as semantic ambiguity and physical implausibility are hardly visible in the pixel space}. To address this, we further incorporate semantic-level textual features of images. \Huang{Following (P3), the TEM converts the image into a caption that serves as explicit semantic evidence rather than as an auxiliary supervision signal. This caption-as-evidence design differs from captioning-based detection, where caption generation itself is the task~\cite{harness}, and from question-only VQA detection~\cite{antifake}, which receives no textual description of the image content.} The TEM is employed to generate captions for the images. These captions are then concatenated with the questions and serve as multi-modal semantic-level textual features to enhance the detection performance of AI-generated images.

In LHSDet, we leverage \Huang{BLIP-2} to generate a detailed caption for the input image and enhance visual understanding. The Q-Former employs a query-based mechanism for finer-grained cross-modal alignment. Specifically, we use the prompt ``\texttt{A detailed description of this image}" to generate descriptive captions encapsulating visual content in linguistic form. The caption of the image serves as a semantic-level textual feature. 
\begin{equation}
\mathbf{T_{\text{desc}}} = f_{\text{BLIP}}(\mathbf{X}, \mathbf{P}_{\text{prompt}}),
\end{equation}
where $\mathbf{X} \in \mathbb{R}^{C\times H \times W}$ denotes the input image.
\Huang{In our implementation, BLIP-2 is kept frozen and the captions are pre-generated offline once per image, which avoids the computational cost of on-the-fly captioning during training and inference. 
}

The generated caption $\mathbf{T_{\text{desc}}}$ is concatenated with the original question to form the complete question $\mathbf{Q}^\prime$ ``\texttt{Based on the image description: } $`\mathbf{T}_{\text{desc}}\text'$, \texttt{assess the authenticity. Is this image real?}". This combined question $\mathbf{Q}^\prime$ is then fed into the LLM to extract textual features. The textual features are utilized as a multi-modal feature in VQA reasoning. This approach effectively translates visual information into the semantic space of the language model, utilizing the advantages of the query-based cross-modal alignment.

\subsection{Multi-Modal Feature Fusion Module}
\Huang{The MFM converts the fused visual feature and the complete question into a unified token sequence for the LLM.}

\Huang{The unified visual representation $\mathbf{F}_{v}$ integrates the low-level local texture feature and the high-level global perception feature, capturing both high-level visual concepts and fine-grained texture details. It is projected into the LLM's embedding space through a linear transformation with bias, $\mathbf{F}_{{vp}} = \mathbf{W}_{{vp}} \mathbf{F}_{{v}} + \mathbf{b}_{{vp}}$, where $\mathbf{W}_{{vp}} \in \mathbb{R}^{D_h \times D_g}$ is the projection matrix and $D_h$ is the LLM hidden dimension, which ensures dimensional compatibility and semantic alignment between visual and linguistic representations.}

The complete question $\mathbf{Q}^\prime$ is processed through standard language tokenization and embedding procedures.

\begin{equation}
\mathbf{F}_{t} = f_{\text{e}}(\text{Tokenize}(\mathbf{Q}')),
\end{equation}
where the tokenization process $\text{Tokenize}(\cdot)$ converts natural language into discrete tokens, while the embedding function $f_{\text{e}}$ maps these tokens to dense vector representations in the same semantic space as the projected visual features.

\Huang{Since textual and visual features are heterogeneous, they are concatenated along the sequence dimension, $\mathbf{F}_{{mm}} = [\mathbf{F}_{{vp}}; \mathbf{F}_{{t}}]$, to preserve their distinct information while enabling joint attention.}

\Huang{The visual tokens serve as a prefix to the textual tokens, so that the language model attends to both modalities simultaneously during processing.}

\subsection{LLM Reasoning Module: Posterior-Calibrated Answer Inference}
\label{sec:lrm}
\Huang{In the LRM, the unified multi-modal token sequence is fed into the LLM, which treats the visual tokens as a prefix and reasons over the joint image--text embedding via self-attention.} The last-token vector of the final-layer hidden state is taken as the LLM’s consolidated representation of the entire image-text input. The answer is then predicted from the final-token representation. We choose the Phi-3-mini model~\cite{phi3} as the LLM reasoning engine for VQA to strike a strategic balance among performance, efficiency, and deployability. Despite its compact size of 3.8B parameters, it has been demonstrated to achieve performance comparable to much larger models like Mixtral 8×7B~\cite{mix} and GPT-3.5~\cite{gpt} on standardized benchmarks.
 
\Huang{The fused visual feature and the complete question are mapped into aligned visual and textual tokens as described in the MFM, and the concatenated sequence is processed by the LoRA-adapted Phi-3-mini:}
\begin{equation}    \mathbf{H}_{\text{LLM}} = f_{\text{Phi-3}}(\mathbf{F}_{mm}).
\end{equation}

The last hidden state corresponding to the final token, \Huang{$\mathbf{h}_{\text{final}} = \mathbf{H}_{\text{LLM}}[:, -1, :] \in \mathbb{R}^{D_h}$,} is extracted.

Within the transformer decoder of Phi-3-mini, the self-attention mechanism processes this multi-modal sequence, enabling each textual token to attend to all visual tokens. This facilitates deep cross-modal reasoning, allowing the model to ground the textual question in the concrete evidence from the visual features. \Huang{Instead of sampling an answer token autoregressively, we read off the logits assigned to ``\texttt{Yes}'' and ``\texttt{No}'' at the final position and normalize them into a posterior probability.}

\Huang{Formally, let $z_{\text{Yes}}$ and $z_{\text{No}}$ denote the logits assigned to the answer tokens ``\texttt{Yes}'' and ``\texttt{No}'' at the final position. The posterior probability that the image is generated (\textit{i.e.}, label $1$) is defined as
\begin{equation}
p = \frac{\exp(z_{\text{No}})}{\exp(z_{\text{Yes}}) + \exp(z_{\text{No}})}.
\label{eq_prob}
\end{equation}

At inference, the image is predicted as generated if $p>0.5$ and as real otherwise, which is equivalent to selecting the answer token with the higher probability.}
\Huang{We term this design \textit{posterior-calibrated answer scoring}: unlike argmax decoding of the answer token~\cite{antifake}, the normalized posterior yields a continuous confidence score for the generated class, which supports threshold tuning and ranking-based metrics such as average precision.}

\subsection{Training Strategy}
\Huang{SigLIP2 and the Phi-3-mini LLM are} fine-tuned using Low-Rank Adaptation (LoRA)~\cite{lora} to adapt to the AI-generated image detection task. LoRA is a widely-used technique for efficiently fine-tuning pre-trained models on downstream tasks.
Under limited data conditions, full fine-tuning can lead to catastrophic forgetting of pre-trained knowledge. To address this, LoRA posits that weight updates during adaptation lie on a low intrinsic dimension, and introduces trainable low-rank matrices to approximate these updates.
\begin{equation}
\mathbf{W}_{\text{p}}^{\text{LoRA}} = \mathbf{W}_{\text{p}} + \Huang{\frac{\alpha}{r}\mathbf{B}_{\text{lora}}\mathbf{A}_{\text{lora}}},
\end{equation}
where $\mathbf{W}_{\text{p}} \in \mathbb{R}^{d \times k}$ denotes the original pre-trained projection matrix, $\Huang{\mathbf{A}_{\text{lora}} \in \mathbb{R}^{r \times k}}$ and $\Huang{\mathbf{B}_{\text{lora}} \in \mathbb{R}^{d \times r}}$ represent the low-rank decomposition matrices with rank $r \ll \min(d,k)$\Huang{, and $\alpha$ is the LoRA scaling factor}.

This parameter-efficient approach enables effective adaptation while preserving the original pre-trained knowledge.
By optimizing only these additional parameters, LoRA preserves the generalizability of the original model while enabling effective task-specific adaptation. 
The proposed LHSDet uses the Binary Cross-Entropy (BCE) loss function:
\begin{equation}
\label{eq_bce}
\mathcal{L}_{\mathrm{BCE}} = \frac{1}{N}\sum_{i=1}^{N}\Bigl[-y_i\log p_i - (1-y_i)\log(1-p_i)\Bigr],
\end{equation}
where $y_i \in\{0,1\}$ denotes the ground-truth binary label, $p_i$ is the model-posterior probability \Huang{defined in Eq.~\eqref{eq_prob}}, and $y_i$ and $p_i$ respectively represent the label and predicted probability for the $i$-th sample in a mini-batch of size $N$.

\section{Evaluations}
\label{sec:exp}
This section is organized as follows. We first describe the datasets and experimental settings, then present visualizations of intermediate results, and finally provide a comprehensive experimental analysis. The analysis encompasses experiments on generalization detection, robustness testing, and ablation studies.

\subsection{Experimental Settings}

\subsubsection{{Datasets}}
\Qian{
Since most existing benchmark datasets are limited to a low resolution, we constructed a set of high‑resolution benchmark for conducting experiments at $1024\times 1024$ pixel resolution collected from ~\cite{flickr},~\cite{antifake},~\cite{aero}. Additionally, we perform cross‑resolution evaluation on the AIGI‑Now~\cite{aiginow} benchmark dataset at $512\times512$ pixels.}

\textbf{(i) High-resolution benchmark.} 
The real-image dataset Flickr2K~\cite{flickr} consists of $2650$ uncompressed images, each with an approximate resolution of $2040\times1356$ pixels. During data preprocessing, the Flickr2K images are first center-cropped to $1024\times1024$ pixels. For training, validation, and testing, the Flickr images are partitioned into $2000/350/300$ images, respectively. The AI-generated image dataset comprises 2650 images generated by SDv3~\cite{sd3} from the public AntifakePrompt~\cite{antifake} database. Identical to the real-data split, the SDv3 images are divided into 2000/350/300 images for training, validation, and testing. Given that the detection accuracy for GAN-generated images has already reached a very high level~\cite{c2pclip}~\cite{aide}, the datasets of synthetic images used for testing in this paper comprise three categories: diffusion models, autoregressive models, and WildRF. \Huang{The diffusion category includes images synthesized by SDv3~\cite{sd3} (the seen generator used for training) and by six unseen generators: SDXL~\cite{sdxl}, Playground v2.5~\cite{playground}, DALL·E 3~\cite{dalle3}, IF~\cite{if}, MidJourney v5, and MidJourney v6.} For \Huang{the} autoregressive model, we generate the images using  Infinity~\cite{infinity}, which creates image datasets based on prompts that cover natural scenery, futuristic cities, science fiction, human portraits, animals, and historical culture. The WildRF benchmark comprises real-world generated images from popular social networks such as Facebook, Reddit, and Twitter. 
\Qian{Except for the Infinity dataset, which is generated in-house, all test data are sourced from publicly available datasets~\cite{antifake,aero,wildrf}.
}
The AI-generated image datasets used for training adopts SDv3, while all other AI-generated image datasets are 
directly employed for testing to evaluate the generalization capability of the detection method.
During training and testing, the input images including real and generated images are center-cropped to $1024 \times 1024$.

\Qian{\textbf{(ii) AIGI-Now~\cite{aiginow} benchmark.} The benchmark dataset comprises images produced by state‑of‑the‑art commercial image generation models, including Flux‑dev, Flux‑Krea, Flux‑\Huang{Kontext}, Flux‑\Huang{Pro}, GPT‑4o, Jimeng, Kling, MiniMax, and Nano Banana. It consists of paired real and generated images and defines two supervision settings. (a) A pixel artifact subset, which uses VAE reconstruction to produce synthetic images that retain semantic information while modifying the statistical properties at the pixel level. (b) A semantically implausible subset, which introduces content inconsistencies via GPT‑4o and then applies intensive postprocessing to remove superficial visual signals. \Huang{For training, a mix of the pixel and semantic subsets of Flux-dev is adopted, including 2,000 real and 2,000 synthetic images. For testing, the pixel and semantic subsets of each remaining generator (Flux-Krea, Flux-Kontext, Flux-Pro, GPT-4o, Jimeng, Kling, MiniMax, and Nano Banana) are used directly without any training; each subset contains 1,000 real and 1,000 generated images. Note that for the $512\times512$ inputs in this cross-resolution evaluation, the local texture block still uses $256\times256$ patches, yielding $2\times2$ patches per image.} During training and testing, the input images are resized and center-cropped to $512 \times 512$.}


















\subsubsection{{Implementation Details}}
The rank-decomposition configurations are set as follows: SigLIP2 and Phi-3-mini \Huang{both} use rank $r= 8$ with LoRA scaling $\alpha= 16$ and dropout $d=0.05$. The target layers for LoRA application are $q_{\text{proj}}$ and $v_{\text{proj}}$. LHSDet is optimized for $\Huang{2}$ \Huang{epochs} using the Adam optimizer with a learning rate of $1 \times 10^{-5}$, weight decay of $0.001$, and a batch size of $1$.
All experiments are performed on a single NVIDIA A100 GPU, and the validation and testing phase uses the same hyperparameters as training.
\Qian{When captions are pre-generated by BLIP-2 and subsequently used for LHSDet training, the proposed LHSDet model \Huang{consists of SigLIP2-So400m (4.4B), Phi-3-mini (3.8B), and approximately 0.2B parameters from the additional low-level visual layers and LoRA adapters, totaling about 8.4B parameters; with 4-bit quantization of Phi-3-mini, the deployment memory footprint is reduced to the equivalent of about 6.6B parameters. During offline caption pre-generation, BLIP-2 OPT-2.7B introduces additional parameters (about 2B-parameter-equivalent memory under 4-bit quantization).}} 
\Qian{The captions are generated by BLIP-2 using a sampling-based beam search with a beam size of 6, temperature 0.9, top-p 0.95, and a maximum length of 60 tokens. Using the pre-generated captions (about 2365 ms per image), LHSDet takes approximately 174.6 ms to process a single $1024\times1024$ image. Remarkably, only $\Huang{2}$ epochs of training are sufficient to achieve high detection performance. }

Following prior work, we employ detection accuracy (ACC) and average precision (AP) as the evaluation metrics. The threshold for accuracy evaluation is set to $0.5$. ACC is calculated as the ratio of correctly classified real and generated images to the total number of real and generated images. AP is defined as the weighted average of precision over all thresholds, weighted by the incremental gain in recall. Since generated images are assigned label 1, AP measures the model's sorting performance for the generated class.

\subsubsection{{Baselines}}
The baselines adopted for comparison include representative methods from recent years, encompassing those based on CNN, RE, and VLM.
\textbf{CNN-based methods:} 
\underline{F3Net} ~\cite{f3net} employs Frequency-aware Image Decomposition to \Huang{extract} forgery patterns.
\Qian{\underline{NPR} ~\cite{upsampling} introduces artifact representations that focus on local up-sampling cues in image pixels.
\underline{SAFE}~\cite{safe} applies three simple image transformations to reduce training bias and enhance local feature awareness.}
\textbf{RE-based methods:} \underline{DIRE}~\cite{dire} calculates the error between an input image and its reconstructed image using DDIM.
\underline{AEROBLADE}~\cite{aero} is a training-free AI-generated image detector using autoencoder reconstruction errors. 
\textbf{VLM-based methods:} \underline{UnivFD} ~\cite{univfd} extracts features using the visual encoder of CLIP ViT\Huang{, and computes} the cosine distance between the input image and the feature spaces of real and generated images. \underline{FatFormer}~\cite{fatformer} is composed of two pre-trained encoders for both image and text prompts with a forgery-aware adapter and language-guided alignment. \underline{AntifakePrompt} ~\cite{antifake} uses InstructBLIP and prompt tuning techniques.
\Qian{\underline{AIDE}~\cite{aide} employs SRM and  CLIP for low-level and high-level feature analysis.}
\Qian{\underline{AlignGemini}~\cite{aiginow} is a two-branch detector combining a semantic VLM and a pixel-artifact expert via logical disjunction.}

\subsection{Visualizing the Intermediate Results}
This section provides visualizations of the intermediate features from the Visual and Textual Encoding Modules of LHSDet, illustrating the interpretability and detection effectiveness of the proposed method.

\subsubsection{Heatmaps of the VEM}
\begin{figure*}
    \centering
       \subfigure[Real image $I_1$.]{
        \includegraphics[width=0.45\linewidth]{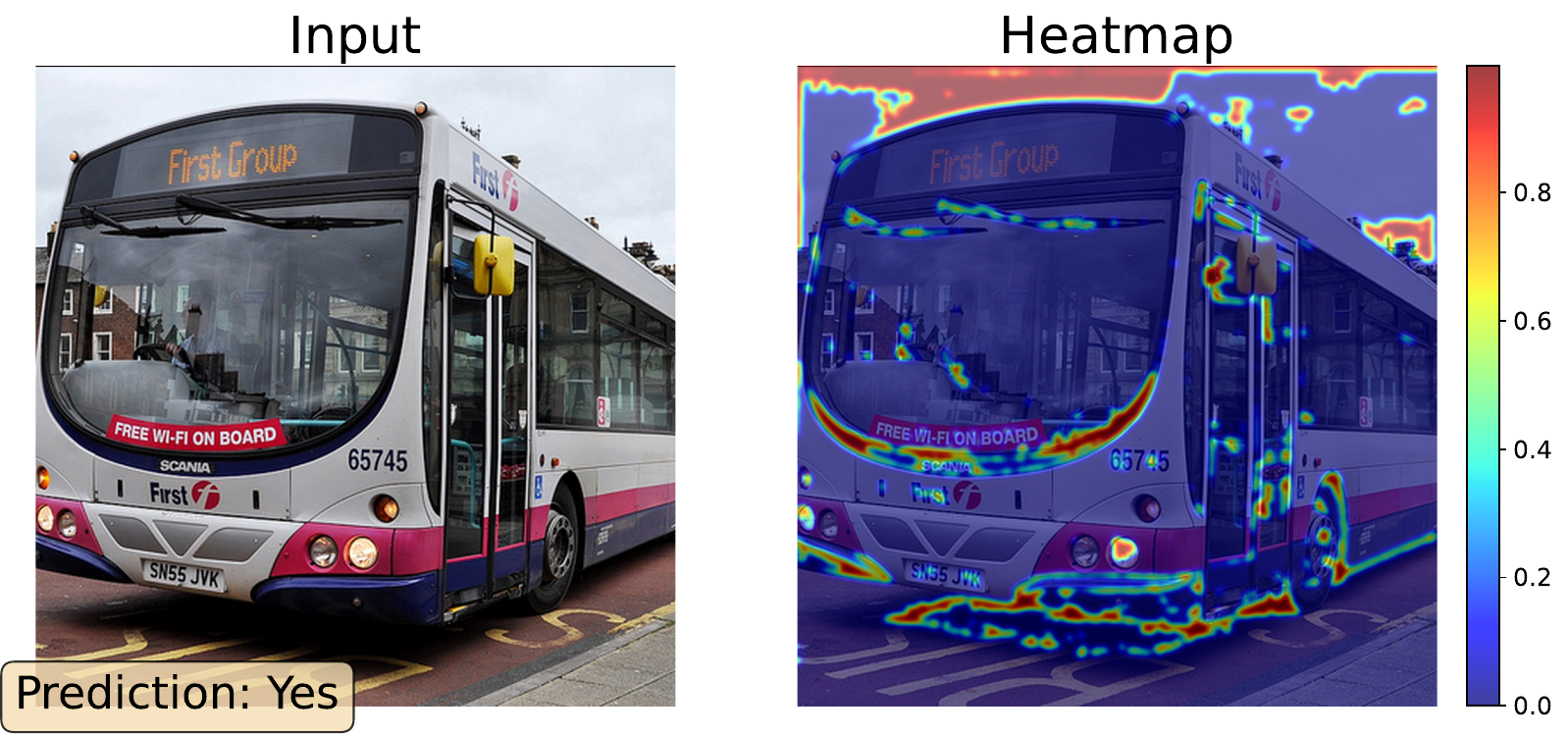}
        \label{fig_flickr}
    }
    \subfigure[AI-generated image $\tilde{I}_1$ by GPT4o (Pixel-level inconsistency).]{
        \includegraphics[width=0.45\linewidth]{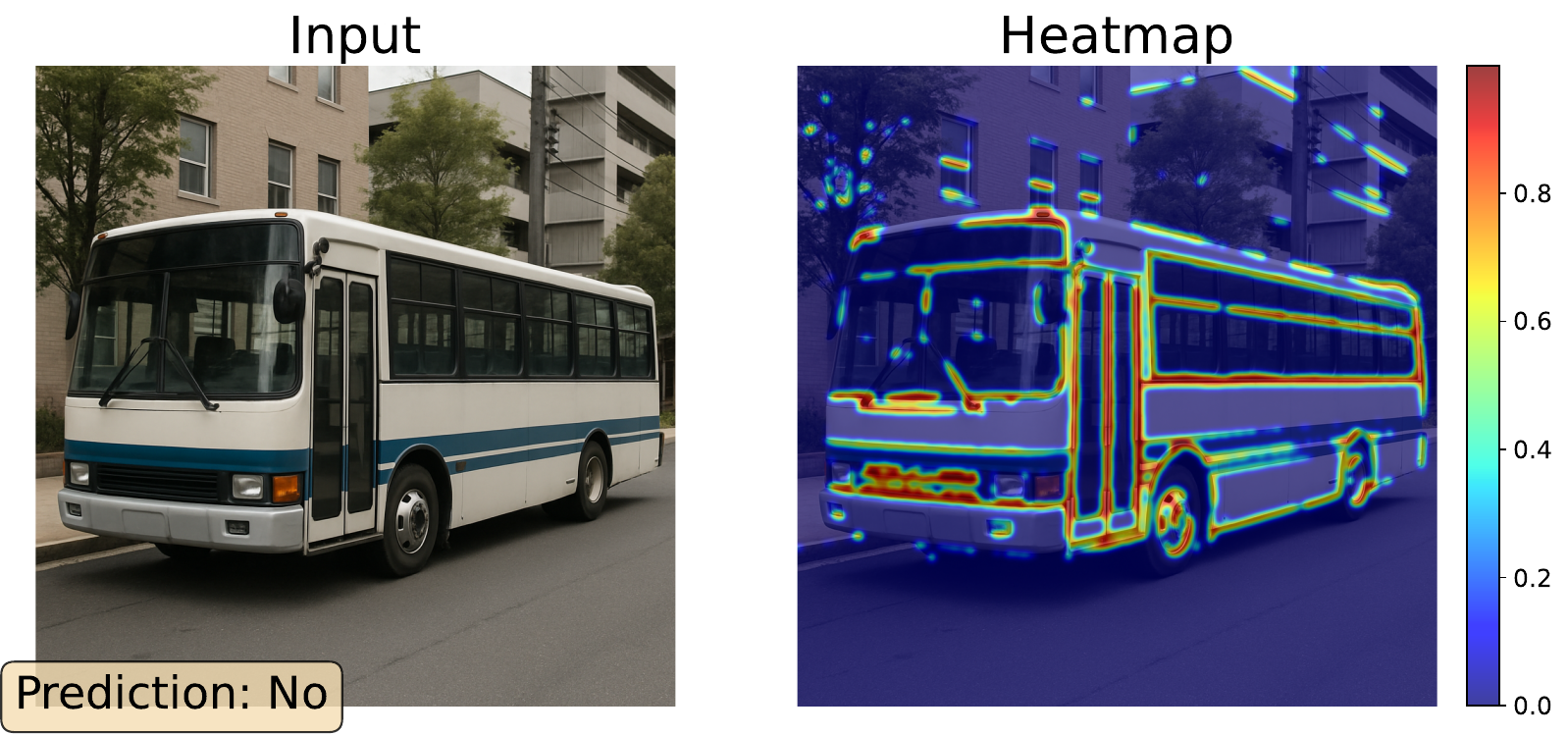}
        \label{fig_sd3}
    }
        \subfigure[Real image $I_2$.]{
        \includegraphics[width=0.45\linewidth]{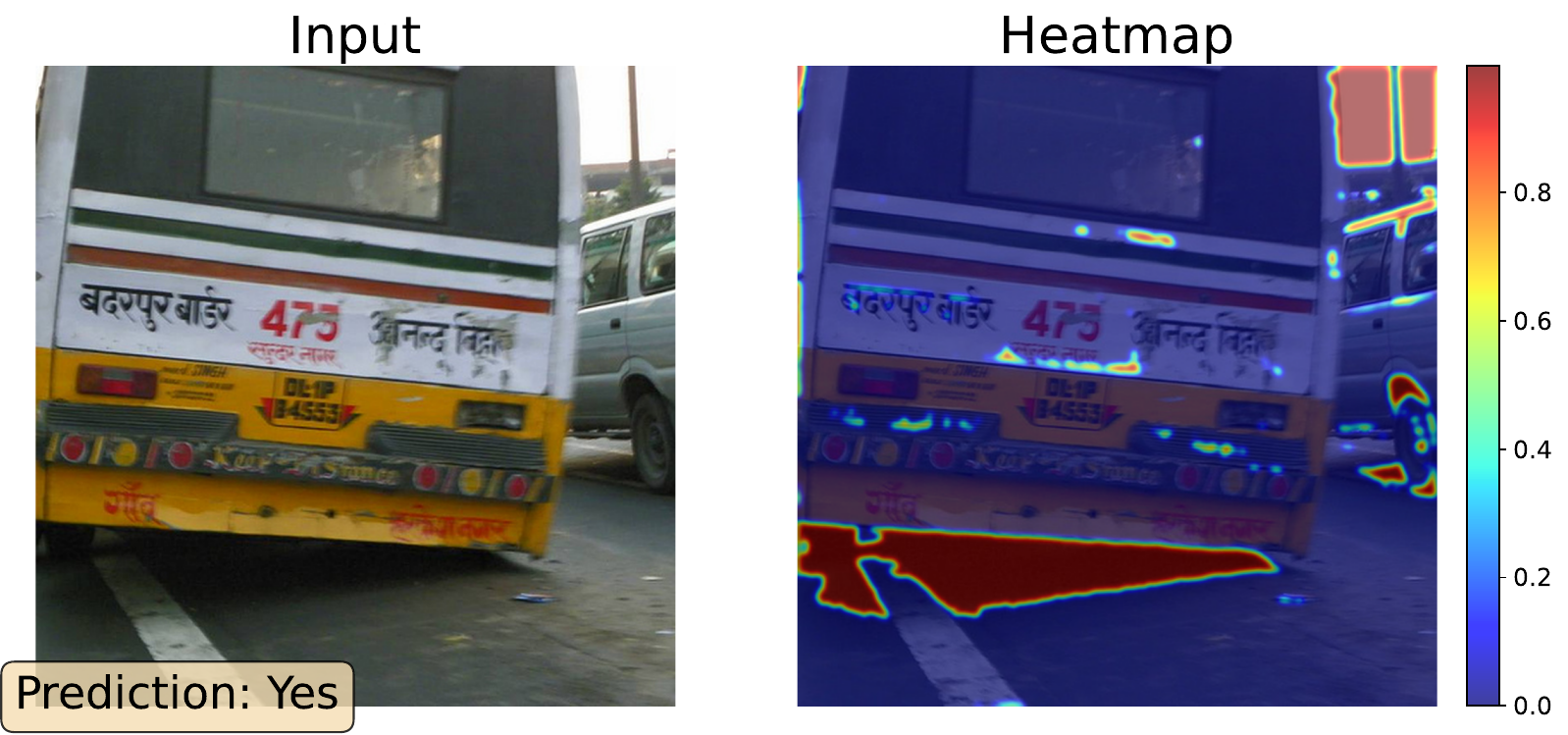}
        \label{fig_div}
    }
    \subfigure[AI-generated image $\tilde{I}_2$ by GPT4o (Semantic-level inconsistency).]{
        \includegraphics[width=0.45\linewidth]{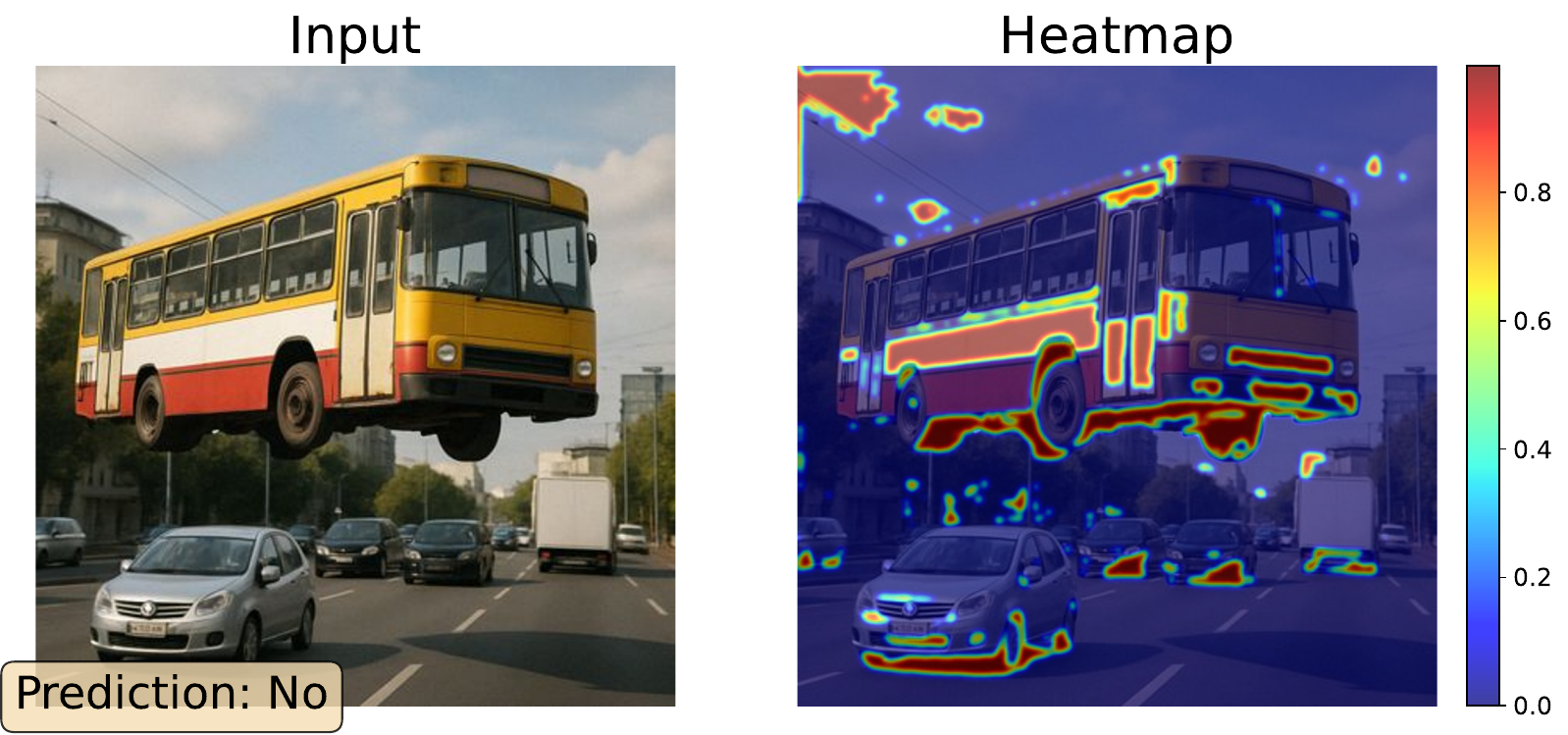}
        \label{fig_infinity}
    }

    \caption{
    \Qian{Visualization of the heatmaps produced by the Visual Encoding Module on real images from Flickr2K and AI-generated images from AIGI-Now (GPT-4o). The heatmap highlights the regions that most influence the detection decision, with red indicating the highest contribution and blue the lowest. 
    }
    }
    \label{fig_heatmaps}
\end{figure*}

\Qian{Specifically, the heatmap is computed as 
$\mathbf{M} = \mathcal{N}(\mathbf{A}_{\text{cross}} \odot \mathbf{A}_{\text{act}})$, where $\mathbf{A}_{\text{cross}}$ denotes the cross-attention weight map produced by the fusion module. It is obtained by projecting the global feature into queries and the local feature map into keys, computing their dot-product similarity at each spatial location, applying softmax over the spatial dimension, and averaging over attention heads. The resulting spatial attention map is then bilinearly upsampled to the input image resolution. 
$\mathbf{A}_{\text{act}}$ is the channel-wise average of the local activation map, also bilinearly upsampled to the same resolution and min--max normalized to 
$[0,1]$. The two maps are multiplied element-wise to obtain the fused heatmap. Finally, the fused heatmap is min--max normalized and further post-processed by thresholding the top 10\%, dilation, distance-weighted expansion, and Gaussian smoothing for visualization.}
\Qian{The resulting heatmap visualizes how cross-attention weights guide local activations during global–local fusion, providing insight into the visual evidence underlying the model’s detection decision.}
The visualization of the heatmaps \Huang{is} shown in \Huang{Fig.}~\ref{fig_heatmaps}. \Qian{It can be observed that in the generated images, there are inconsistencies in the high-frequency details of textures and edges, as well as semantic irregularities in fine-grained features such as the abnormal ground clearance of the bus. In contrast, the real images exhibit natural transitions in texture and lighting details and normal semantic characteristics.}

\subsubsection{Caption Feature of the TEM}
\begin{figure*}[htbp]
    \centering
    \subfigure[The AI-generated images and their captions.]{
        \includegraphics[width=0.45\linewidth]{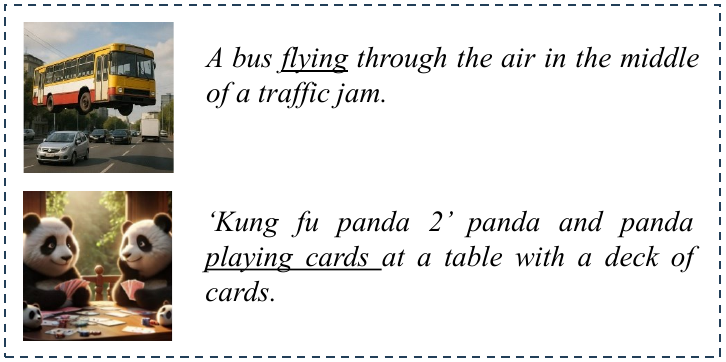}
        \label{fig_caption_gen}
    }
    \subfigure[The real images and their captions.]{
        \includegraphics[width=0.45\linewidth]{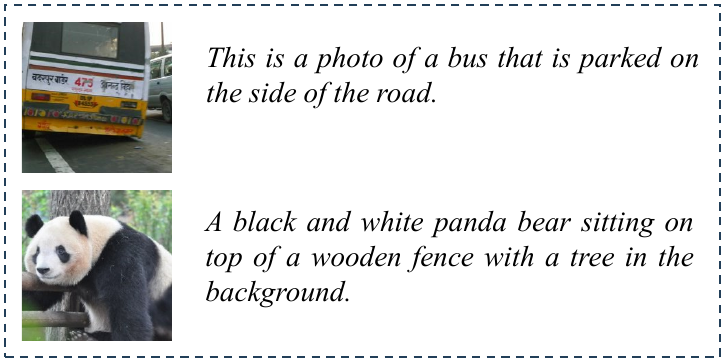}
        \label{fig_caption_real}
    }
    \caption{
    \Qian{Visualization of the caption features generated by BLIP-2. Captions of AI-generated images (a) often describe physically implausible or logically contradictory content (\textit{e.g.}, "a bus flying through the air"), whereas captions of real images (b) are consistent with natural laws. These semantic cues provide complementary evidence for detection beyond visual features.}
    }
    \label{fig_caption}
\end{figure*}

The visualization of the caption extraction of the input image is shown in Fig.~\ref{fig_caption}. It can be observed that physically implausible or logically contradictory elements are present in some generated image captions, whereas captions of real images adhere to natural laws. This caption-based semantic representation offers complementary textual clues for AI-generated image detection, rather than relying purely on visual feature analysis.

\begin{table*}[t]
\setlength{\tabcolsep}{3.5pt}
\caption{
\Huang{Detection accuracy (ACC, \%) on the high-resolution benchmark. All methods are trained on Flickr2K (real) and SDv3 (generated) at $1024\times1024$ resolution and evaluated on eleven test sets covering diffusion models, an autoregressive (AR) model, and the WildRF benchmark. (The best and the second best results in each column are in bold and underlined, respectively.) Min ACC denotes the minimum accuracy across the ten unseen test sets.}
}
\begin{center}
\begin{threeparttable}
\begin{tabular}{l|ccccccc|c|ccc|cc} 
\toprule
\multirow{2}{*}{\textbf{Methods}} &
\multicolumn{7}{c}{\textbf{Diffusion Models}} &
\multicolumn{1}{|c}{\textbf{AR Model}} &
\multicolumn{3}{|c|}{\textbf{WildRF}} &
\multirow{2}{*}{\textbf{Avg ACC}} &
\multirow{2}{*}{\Huang{\textbf{Min ACC}}} \\
\cmidrule{2-9}\cmidrule{10-12}
& \textbf{SDv3} & \textbf{SDXL} & \textbf{Playground} & \textbf{DALL·E3} &
\textbf{MJv5} & \textbf{MJv6} & \textbf{IF} &
\textbf{Infinity} &
\textbf{Facebook} & \textbf{Reddit} & \textbf{Twitter}
\\ 
\midrule


F3Net~\cite{f3net}&95.83 	&52.67 	&54.83 	&60.50 	&59.83 	&56.33 	&63.50 	&65.00	&55.67 	&55.00 	&55.83 	&61.36 &\Huang{52.67} \\


\Qian{NPR}~\cite{upsampling} &75.50 & 	58.00 &	56.50 	&68.33 	&60.00 &	55.83 &	74.33 &	54.00 &	59.33 &	60.83 &	60.83 	&62.13 &\Huang{54.00} \\

\Qian{SAFE}\cite{safe}&	69.00 &	71.17 	&56.17 &	64.67 &	64.83 &	54.00 &	60.83 	&69.00 &	55.00 	&56.00 &	57.50& 	61.65 &\Huang{54.00} \\

DIRE~\cite{dire} &{98.26}  &72.40 &66.50  &86.22 	 &69.05  &69.22  &	93.71  & 74.00 &78.31 &65.99&66.33	&76.36 &\Huang{65.99} \\

AEROBLADE~\cite{aero} &83.96 	&47.77 &	85.90 &	\textbf{99.42} &	\underline{94.36} &	\textbf{91.48} &	91.47 &	\underline{85.02} &	86.03 &	\underline{70.71} &	\underline{82.21} &	83.48 
&\Huang{47.77}\\

\Qian{UnivFD}~\cite{univfd} &96.33 &	\underline{92.67} &	\underline{93.33} &	87.17 &	82.50 &	77.33 &	96.33 &84.00 &\underline{90.00} &	70.33 &	82.17 &\underline{86.56} &\Huang{\underline{70.33}} \\
 
\Qian{FatFormer}~\cite{fatformer} &\underline{98.50} &69.17 &81.50 &90.17 &	79.83 &	62.67 &\underline{96.83} &82.00 &77.67 &	63.83 &	71.67 &	79.44 &\Huang{62.67} \\

Antifake~\cite{antifake} &93.33 &{90.17} &{92.50} &80.17 &79.67 &73.83 &86.33 &84.00 &80.00 &68.00 &73.83 &81.98 &\Huang{68.00} \\

\Qian{AIDE}~\cite{aide} &97.83 &55.83 &52.17 	&69.00 &59.17 &57.33 &75.50 &52.00 &52.67 &54.83 &52.67 &	61.73 &\Huang{52.00}  \\

\Qian{AlignGemini}$\dagger$~\cite{aiginow} &- &- &- &- &- &-&- &- &\underline{92.20} &\underline{90.90} &\textbf{95.50} &- &-\\

\midrule
\textbf{LHSDet}  (ours) & \textbf{99.83} 	&\textbf{98.83} 	&\textbf{99.17} 	&\underline{97.83} 	&\textbf{94.50} 	&\underline{85.33} 	&\textbf{99.00} 	&\textbf{99.00} 	&\textbf{98.00} 	&\textbf{93.67} 	&\underline{95.33} 	&\textbf{96.41} &\Huang{\textbf{85.33}}
\\

\bottomrule
\end{tabular}
\end{threeparttable}
\end{center}
\label{tab_fewshot_flickr}
\vspace{-0.5cm}
\end{table*}

\subsection{Accuracy of Generalization Detection}
In practical scenarios, the emergence of unknown generative models makes extensive pre-training datasets inaccessible. To address this challenge, our experiment employs fewer samples for both training and evaluation. The comparison of detection results on the Flickr2K and $11$ generated-image datasets are shown in Table~\ref{tab_fewshot_flickr}, the training dataset comprises $2,000$ generated and $2,000$ real images respectively, while the test set contains no more than $300$ samples per category. Furthermore, substantial architectural discrepancies exist across different generative models. And generated images from various sources are often widely spread through online platforms. To better approximate real-world conditions, we conduct cross-model generalization tests incorporating three distinct categories: diffusion models, autoregressive models, and the WildRF benchmark~\cite{wildrf} (generated images obtained from social media). This comprehensive evaluation ensures our assessment reflects the diverse generative artifacts encountered in practical applications. 

In Table~\ref{tab_fewshot_flickr}, we can observe that the CNN-based detection methods such as F3Net, \Qian{NPR, and SAFE} suffer a sharp drop in generalization, with detection accuracies of only $61\%-62\%$. The RE-based methods achieve better detection performance than the CNN-based detectors. DIRE attains an accuracy of approximately $76.36\%$, yet its results remain unsatisfactory on most datasets. AEROBLADE, a training-free approach,
delivers overall better performance with an accuracy of approximately $83.48\%$. Nevertheless, it performs poorly on the SDXL and WildRF datasets. 
\Qian{In experiments on VLM-based detectors, the CLIP-based method UnivFD achieves an accuracy of approximately $86.56\%$, while the detection capability of FatFormer and AIDE remains unsatisfactory at $61\%-79\%$, which exhibits limited generalization.}
The BLIP-based detector AntifakePrompt achieves an overall accuracy of approximately $81.98\%$. However, it only performs effectively on a limited number of datasets, while its detection accuracy on Midjourney and WildRF remains unsatisfactory. AlignGemini achieves competitive performance on the WildRF dataset. Our method achieves the best detection performance, attaining an accuracy of $96.41\%$ across three categories and eleven datasets, substantially surpassing all baselines by $9.85\%$ over the second-best baseline ($86.56\%$). \Huang{Moreover, in terms of the worst-case accuracy across the ten unseen test sets (Min ACC in Table~\ref{tab_fewshot_flickr}), LHSDet achieves 85.33\%, exceeding the best baseline (UnivFD, 70.33\%) by 15 percentage points. This indicates that LHSDet remains reliable even against the most challenging unknown generator, which is critical for practical deployment.} \Huang{Note that each test subset contains no more than 300 images, for which the 95\% confidence interval of ACC is about $\pm1.6\%$ near the ceiling; differences within this range should be interpreted with caution. 
} 
This gain is attributed to our triple-branch architecture, multi-modal feature fusion, and LoRA fine-tuning.

\Qian{To validate the effectiveness of LHSDet under cross-resolution and downsampled inputs, we conduct a \Huang{cross-resolution evaluation} with inputs downsampled to $512\times512$. Based on Tables~\ref{tab_pixel} and ~\ref{tab_semantic}, we conduct cross-resolution evaluations on the AIGI-Now benchmark, separately for the pixel-artifact-discriminative and semantic-discriminative subsets. This benchmark deliberately applies aggressive post-processing to disrupt pixel-level statistics and disentangle semantic cues from pixel-level cues
\Qian{making pixel-level clues more difficult to detect. As shown in Table~\ref{tab_pixel}, existing VLM-based detection methods all exhibit significant performance degradation on this subset. Reconstruction-based methods DIRE and AEROBLADE rely on pixel-level reconstruction errors. When these cues are perturbed, their detection performance shows a notable decline, particularly that of AEROBLADE, which drops by approximately 20\% compared with the high-resolution benchmark. For the semantic-discriminative subset, CNN-based detectors such as F3Net, NPR, and SAFE exhibit poor performance on both the pixel and semantic subsets, with overall average ACC and AP roughly between 52\%–68\% and 52\%–77\%, indicating limited generalization. 
}
VLM-based methods such as UnivFD, FatFormer, Antifake, achieve near-perfect performance on the semantic-discriminative subset, around 96\%–97\%, demonstrating the clear advantage of VLM-based discrimination that leverages global semantic clues.
However, they suffer significant drops on the pixel-artifact-discriminative subset, around 85\%–87\%.
AlignGemini achieves 92.14\% accuracy on the pixel subset, showing a modest advantage, but its semantic-subset accuracy is about 93.18\%, a slight drop compared with other VLM-based methods.
In contrast, our LHSDet ranks first in average ACC on both subsets, with AP essentially tied with FatFormer on the semantic subset.
On the semantic subset, its average ACC and AP are 98.13\%/99.91\%, and it also ranks first on the pixel subset with average ACC and AP of 94.33\%/98.88\%, clearly demonstrating its superior generalization capability against diverse AI‑generated imagery. Our method demonstrates a clear advantage over existing VLM‑based detectors, particularly in image detection on the pixel subset. This indicates that default VLM detectors are not equipped with low‑level artifact detection capabilities, whereas the low‑level branch of LHSDet substantially enhances the capture of texture details in AI‑generated images.}

\begin{table*}[t]
\caption{
\Huang{Detection accuracy and average precision (ACC/AP, \%) on the \textbf{pixel-artifact-discriminative subset} of AIGI-Now at $512\times512$ resolution. All methods are trained on Flux-dev (a mixture of its pixel and semantic subsets) and evaluated on the remaining eight generators without further training. (The best and second-best results are shown in bold and underlined, respectively.) $\dagger$ indicates that the results are obtained from~\cite{aiginow}}.
}
\begin{center}
\begin{threeparttable}
\begin{tabular}{l|cccccccc|c}
\toprule
\textbf{Methods} 
& \textbf{Flux-Krea}	& \textbf{Flux-Kontext} &\textbf{ Flux-Pro}	& \textbf{GPT4o}	& \textbf{Jimeng} & \textbf{Kling}	 & \textbf{MiniMax}	& \textbf{Nano Banana}& \textbf{ Avg ACC/AP}	
 \\ 
\midrule
\Qian{F3Net}~\cite{f3net} &	68.20/70.22 &	59.00/62.35 &	47.30/49.30 &	54.20/57.77 &	53.75/52.94 &	63.10/64.45 &	63.80/64.98& 	57.80/60.88  & 58.39/60.36 \\

\Qian{NPR}~\cite{upsampling} &	54.70/53.98 &	56.05/59.77  &	55.35/54.91 & 55.05/51.33 & 49.45/47.06 & 52.80/50.56 & 53.10/53.89 & 55.45/49.40 & 53.99/52.61 \\


\Qian{SAFE}~\cite{safe}  &	55.00/53.39  & 52.50/50.92  & 47.05/45.16  &55.10/61.12  & 43.70/43.22  &55.15/59.01  &51.80/49.37  & 56.95/64.84  &	52.16/53.38 \\

\Qian{DIRE}~\cite{dire} &77.95/91.09 &	66.30/76.83 &	62.00/75.46 	&76.95/89.60 	&68.40/83.54 	&84.75/93.39 	&77.35/88.47 &	73.35/86.33 &	73.38/85.59  \\

\Qian{AEROBLADE}~\cite{aero} &66.63/70.44 	&56.83/58.41 &	58.85/61.28 &	68.63/69.71  &67.85/73.10 
&64.21/66.80 	& 60.71/63.94  &56.60/56.82 	&62.54/65.06  \\

\Qian{UnivFD}~\cite{univfd}  &\underline{94.20}/\underline{98.64}  &65.65/80.85  &79.90/92.00  &{91.95}/97.53  &\underline{94.75}/\underline{98.89} 	&\underline{93.95}/\underline{98.66}  &86.45/94.96  &{95.30}/\underline{99.11} 	 &{87.77}/{95.08} \\

\Qian{FatFormer}~\cite{fatformer} & 89.50/97.12  &{75.00}/\underline{85.96}  &{86.95}/\underline{97.13}  & 87.50/\underline{98.88}  &87.30/98.80  &87.00/97.63 & {87.10}/\underline{97.59}  &88.15/96.84 	 & 86.06/\underline{96.24} \\

\Qian{Antifake}~\cite{antifake} & 92.05/97.29  &	67.30/75.86  & 78.95/87.74  &91.05/96.91 	& 89.60/96.69  & 87.95/96.08 & 83.25/91.92  & 89.95/95.73 	 & 85.01/92.28 \\

\Qian{AIDE}~\cite{aide} &64.60/80.73 	&58.45/69.65 	&61.55/74.27 	&65.15/80.22 &	78.25/89.46 	&66.40/80.70 &	62.15/73.22 	&58.60/73.72 	&64.39/77.75 \\

\Qian{AlignGemini}$\dagger$~\cite{aiginow} &	83.00/- 	&	\textbf{86.60}/- &\underline{89.90}/-	&	\underline{96.80}/- 	&	\textbf{96.20}/- 	&\underline{97.30}/- 	&	\underline{91.30}/-	&	\underline{96.00}/-		&\underline{92.14}/- \\ 	

\midrule									
\textbf{\Qian{LHSDet} }  (ours) &\textbf{96.00}/\textbf{99.66}  &\underline{83.90}/\textbf{93.72}  &\textbf{94.65}/\textbf{99.51} 	 &\textbf{98.45}/\textbf{99.89}&{92.95}/\textbf{99.24}  &\textbf{97.45}/\textbf{99.86} 	&\textbf{92.10}/\textbf{99.20} &\textbf{99.15}/\textbf{99.98}  &\textbf{94.33}/\textbf{98.88} \\

\bottomrule

\end{tabular}
\end{threeparttable}
\end{center}
\label{tab_pixel}
\end{table*}

\begin{table*}[t]
\caption{
\Huang{Detection accuracy and average precision (ACC/AP, \%) on the \textbf{semantic-discriminative subset} of AIGI-Now at $512\times512$ resolution. Training and evaluation protocols follow Table III. (The best and second-best results are shown in bold and underlined, respectively.) $\dagger$ indicates that the results are obtained from~\cite{aiginow}.}
}
\begin{center}
\begin{threeparttable}
\begin{tabular}{l|cccccccc|c}
\toprule
\textbf{Methods} & \textbf{Flux-Krea}	& \textbf{Flux-Kontext} &\textbf{ Flux-Pro}	& \textbf{GPT4o}	& \textbf{Jimeng} & \textbf{Kling}	 & \textbf{MiniMax}	& \textbf{Nano Banana}& \textbf{ Avg ACC/AP }	
 \\ 
\midrule

\Qian{F3Net}~\cite{f3net} &	64.75/72.25 &	61.10/68.65 &	52.50/60.59 &	48.75/57.50 &	53.40/55.85 &	55.15/62.58 &	54.95/63.51 &	48.75/58.59 &	54.92/62.44 \\

\Qian{NPR}~\cite{upsampling}  &56.75/64.00 &55.85/65.16 &	57.85/60.22 &	50.25/47.91 &49.20/45.10 &47.95/46.17 &	54.85/56.35& 	55.00/48.87 & 53.46/54.22 \\


\Qian{SAFE}~\cite{safe} & 68.90/78.42 & 69.50/79.97 &	71.20/82.34 &	66.85/75.69 & 63.55/73.57 & 66.90/73.51 & 68.95/77.07 &	68.45/75.29 &	68.04/76.98 \\

\Qian{DIRE}~\cite{dire} &80.50/92.86 &74.30/84.88 &85.90/93.93 &76.70/89.33 &84.95/93.86 &86.10/94.41 &80.60/90.74 &80.40/92.45 &81.18/91.56   \\

\Qian{AEROBLADE}~\cite{aero}  & 84.43/89.30 & 61.01/65.80 & 52.75/52.05 & 61.90/62.20  & 60.78/64.89 & 62.13/67.36 
 & 54.24/54.87 &	55.48/47.13  & 61.59/62.95  \\

\Qian{UnivFD}~\cite{univfd} &\underline{98.45}/99.90 &	85.55/96.93 &97.85/99.66  &\underline{99.05}/99.93 & \underline{99.25}/99.94 & 98.80/99.88 & 96.80/\underline{99.60}  &\underline{99.15}/99.96 &	96.86/99.48 \\

\Qian{FatFormer}~\cite{fatformer} & 97.80/\underline{99.96} & \textbf{92.10}/\textbf{99.95} &	\underline{98.85}/\textbf{99.99}	& 97.90/\underline{99.97} & 98.45/\underline{99.98} & \underline{99.10}/\textbf{99.99} &	\underline{97.05}/\textbf{99.98} & 97.70/\underline{99.98} & \underline{97.37}/\textbf{99.98} \\

\Qian{Antifake}~\cite{antifake} &	98.25/99.89 & 84.10/95.82 &	97.95/99.70 & 98.75/99.88 & 98.85/99.90 &	97.85/99.77 &	96.85/99.51 &	98.65/99.93 &	96.41/99.30 	\\

\Qian{AIDE}~\cite{aide} &80.30/94.30 &	74.30/90.82 	&72.70/88.69 	&73.95/90.52 &	84.30/95.31 	&74.40/89.51 	&67.30/84.33 &	69.45/89.63 	&74.59/90.39 \\

\Qian{AlignGemini}$\dagger$~\cite{aiginow} &92.60/-  &88.00/- &93.20/- 	&95.70/- 	&95.60/- 	&94.90/- &89.80/- 	&95.60/- 	&93.18/- \\

\midrule
\textbf{\Qian{LHSDet}}	 (ours)
&\textbf{99.10}/\textbf{99.99} & \underline{90.80}/\underline{99.33} &	\textbf{98.90}/\underline{99.97} &	\textbf{99.45}/\textbf{100}	& \textbf{99.95}/\textbf{100}	& \textbf{99.40}/\underline{99.98} &	\textbf{97.50}/\textbf{99.98} &	\textbf{99.95}/\textbf{100}	& \textbf{98.13}/\underline{99.91} \\

\bottomrule

\end{tabular}
\end{threeparttable}
\end{center}
\label{tab_semantic}
\end{table*}

\begin{table*}[t]
\caption{Robustness against unseen post-processing procedures of different AI-generated image detectors on the Flickr2K and SDv3 datasets with respect to Accuracy (ACC, \%). (The best and the second best results in each column are in bold and underlined.) 
}
\begin{center}
\begin{threeparttable}
\begin{tabular}{l|ccccccc|cc}
\toprule
\textbf{Methods} &\textbf{Clear} &\textbf{Brightness} &\textbf{Contrast} &\textbf{Gaussian Blur} &\textbf{JPEG} &\textbf{Gaussian Noise} &\textbf{Rotation} &\textbf{Avg ACC$\uparrow$} &\textbf{Avg Drop$\downarrow$}\\
\midrule

F3Net~\cite{f3net} &95.83 &	90.17 &	90.83 &	55.17 &54.67 &54.00 &63.17 &68.00 &27.83  \\


\Qian{NPR}~\cite{upsampling} &75.50 &68.17 &	68.00 &	75.50 &	74.33 &	73.83 &	67.33 &71.19 &4.31 \\

\Qian{SAFE}~\cite{safe} &69.00 &67.83 &67.83 &	67.67 &	67.83 &	61.33 &	56.33 &64.80 &4.20 \\

DIRE~\cite{dire} &98.26 &95.49  & \underline{97.22} & 95.66 & \underline{98.31} &\underline{95.14} &\textbf{98.44} &\underline{96.71} &\underline{1.55}  \\

AEROBLADE~\cite{aero} &83.96 &	67.32 	&64.66 &	79.31 &	77.11 &	65.66 	&49.68 &	67.29 &	16.67  \\

\Qian{UnivFD}~\cite{univfd} &	96.33 &	91.50 &	90.67 &	96.17 &	96.17 &	92.83 	&93.17 	&93.42 & 2.91 \\

\Qian{FatFormer}~\cite{fatformer} &	\underline{98.50} 	&\underline{97.33} &96.50 &\underline{98.00}  &97.50 &92.00 	&\underline{98.33} 	&96.61  &1.89 \\

Antifake~\cite{antifake} &93.33 &93.17 &92.83 &	91.67 &91.17 &91.17 	&93.33 &92.22 &\textbf{1.11} \\

\Qian{AIDE}~\cite{aide} &97.83 	&92.83 	&93.83 &	97.67 &	97.83 	&\textbf{97.50}& 	61.50 &	90.19 &	7.64 
 \\

\midrule
\textbf{\Qian{LHSDet}}  (ours) &\textbf{99.83} &\textbf{99.17} 	&\textbf{98.83} &\textbf{99.17}	&\textbf{99.83} &94.83 	&96.50 	&\textbf{98.06} &1.78  \\

\bottomrule
\end{tabular}
\end{threeparttable}
\label{tab_robustness_sd3}
\end{center}
\end{table*}

\subsection{Robustness against unseen Post-Processing Procedures}
In this experiment, the robustness testing against unseen image post-processing procedures includes: brightness adjustment, contrast modification, Gaussian blur (kernel \Huang{size} of $3\times3$ and \Huang{$\sigma_{\text{blur}}=1$}), JPEG compression (quality \Huang{factor} $Q_{\text{JPEG}}=85$), random Gaussian noise (\Huang{$\sigma_{\text{noise}}$ uniformly sampled from} $[5, 20]$), and \Huang{random rotation with $\theta$ uniformly sampled from} $[-90^{\circ}, 90^{\circ}]$. 
The comparisons for robustness against unseen post-processing procedures on the Flickr2K and diffusion model SDv3 where each dataset consists of \Qian{no more than} $300$ test samples, as seen in Table~\ref{tab_robustness_sd3}.
\Qian{AntifakePrompt, DIRE, our proposed LHSDet, and Fatformer all demonstrate strong robustness, with average accuracy drops of approximately $1\%-2\%$.
\Huang{Among them, AntifakePrompt achieves the smallest average drop (1.11\%), whereas LHSDet attains the highest absolute accuracy under perturbations (average 98.06\%), offering the best accuracy--robustness trade-off.}
UnivFD achieves a drop of $2.91\%$. 
AIDE suffers from limited robustness to image rotation, with an overall average reduction of $7.64\%$.
Notably, NPR and SAFE exhibit relatively small drops of $4\%$, but their clear image accuracies are only $71.19\%$ and $64.80\%$, which indicates insufficient discriminative capability.}
CNN-based method F3Net shows severe degradation, which suffers an extreme drop of $27.83\%$, indicating high fragility. 
The reconstruction-based method AEROBLADE shows a substantial drop of $16.67\%$, particularly under rotation.

\begin{table*}[t]
\caption{Ablation study on key components.
Accuracy (ACC, \%) of LHSDet and its variants with individual components removed.
}
\begin{center}
\begin{threeparttable}
\begin{tabular}{l|ccccccc|c|ccc|c}
\toprule
\multirow{2}{*}{\textbf{Methods}} &
\multicolumn{7}{c}{\textbf{Diffusion Models}} &
\multicolumn{1}{|c}{\textbf{AR Model}} &
\multicolumn{3}{|c|}{\textbf{WildRF}} &
\multirow{2}{*}{\textbf{Avg ACC}}  \\ \cmidrule{2-9}\cmidrule{10-12}
& \textbf{SDv3} & \textbf{SDXL} & \textbf{PlayGround} & \textbf{DALL·E3} &
\textbf{MJv5} & \textbf{MJv6} & \textbf{IF} &
\textbf{Infinity} &
\textbf{Facebook} & \textbf{Reddit} & \textbf{Twitter} & \\ \midrule

w/o Low-Level &99.67 	&96.33 	&98.83 	&97.17 	&85.00 	&71.50 	&99.00 	&97.00 	&88.00 	&77.50 &83.33 &90.30 \\

w/o High-Level &59.33 	&62.33 	&61.50	&59.33 &54.67 &49.33 	&60.00 &57.00 &60.67 &61.67 &60.17 	&58.73  \\

w/o Semantic-Level &99.67 	&97.67 	&98.50 	&97.83 	&90.83 	&77.00 	&99.33 	&99.00 	&95.33 	&89.67 	&92.33 	&94.29 \\

w/o LoRA &98.33 	&96.83 	&98.17 	&94.83 	&89.33 	&80.50 	&98.67 	&96.00 	&92.67 	&91.83 	&92.17 	&93.58  \\

\Qian{w/o Patch Aggregation} & 97.33 	&97.33 &	96.67 	&96.00 &93.00 &	85.00 &	96.67 &	99.00 &	97.67 	&93.33 &	95.17 &	95.20 \\






\textbf{Full LHSDet} & {99.83} &{98.83} &{99.17} &97.83 &{94.50} &{85.33} &99.00 &{99.00} &{98.00} &{93.67} &{95.33} &\textbf{96.41}  \\

\bottomrule
\end{tabular}
\end{threeparttable}
\label{tab_ablation}
\end{center}
\end{table*}

\begin{table*}[t]
\caption{
\Huang{Ablation study on backbone and fusion-mechanism. Accuracy (ACC, \%) of LHSDet variants with different visual encoders (CLIP, DINOv2, SigLIP2), LLMs (Vicuna-7B, Phi-3-mini), and visual-branch fusion strategies (cross-attention, gated fusion). "Cross-Att." denotes the proposed global-guided cross-attention. All variants follow the training protocol in Section IV-A2.}
}
\begin{center}
\begin{threeparttable}
\begin{tabular}{l|ccccccc|c|ccc|c}
\toprule
\multirow{2}{*}{\textbf{Methods}} &
\multicolumn{7}{c}{\textbf{Diffusion Models}} &
\multicolumn{1}{|c}{\textbf{AR Model}} &
\multicolumn{3}{|c|}{\textbf{WildRF}} &
\multirow{2}{*}{\textbf{Avg ACC}}  \\ \cmidrule{2-9}\cmidrule{10-12}
& \textbf{SDv3} & \textbf{SDXL} & \textbf{PlayG.} & \textbf{DALL·E3} &
\textbf{MJv5} & \textbf{MJv6} & \textbf{IF} &
\textbf{Infinity} &
\textbf{Facebook} & \textbf{Reddit} & \textbf{Twitter} & \\ \midrule






\Qian{CLIP+Vicuna+\Huang{Cross-Att.}} &97.83 	&90.33 	&93.67 &	84.33 &	81.67 &	68.50 &	95.15 &	90.00 &	86.00 &	75.67 	&84.67 &	86.17 \\ 

\Qian{DINO2+Vicuna+\Huang{Cross-Att.}}	&91.50 &	83.50 &	84.67 	&85.33 	&81.92 &	61.67 	&91.33 &	75.00 	&74.00 	&63.83 	&70.33 &	78.46 \\

\Qian{DINO2+Phi3+\Huang{Cross-Att.}}	&88.83 &	84.50& 	88.17 	&82.67 &	76.50 &	69.50 	&88.33 &	76.00 &	79.67 	&63.17 &	74.00 &	79.21 \\

\Qian{SigLIP2+Vicuna+\Huang{Cross-Att.} }&	99.67 &	{99.17} &	{99.33} &	97.17 &	91.67 &	82.67 &	97.50 &	99.00 &	95.67 &	91.33 &	94.33 &	95.23 \\

SigLIP2+Phi3+Gate & 99.33 &98.00 &98.33 &{98.33} &93.00 &80.00 &{99.50} &99.00  &94.67  &93.00 &94.17 &95.21  \\

\textbf{SigLIP2+Phi3+\Huang{Cross-Att.}} & {99.83} &{98.83} &{99.17} &97.83 &{94.50} &{85.33} &99.00 &{99.00} &{98.00} &{93.67} &{95.33} &\textbf{96.41}  \\

\bottomrule
\end{tabular}
\end{threeparttable}
\label{tab_backbone}
\end{center}
\end{table*}

\subsection{Ablation Study}

To quantify the contribution of each component, we conduct systematic component-wise ablation studies targeting the low-, high-, semantic-level feature, and LoRA fine-tuning configuration. 
As demonstrated in Table~\ref{tab_ablation}, each component contributes to the improvement of detection accuracy. Removing either the low-level or the high-level branch leads to a clear degradation in detection accuracy, indicating that both branches are mutually complementary and work collaboratively. Specifically, the absence of the low-level branch causes a sharp decline in detection accuracy for datasets such as MidJourney and WildRF. \Huang{Notably, the contribution of the low-level branch is most pronounced on unseen generators: removing it decreases the accuracy on MJv5, MJv6, and the three WildRF subsets by 9.50, 13.83, and 10.00--16.17 percentage points, respectively, which directly supports the motivation that low-level texture cues are essential for generalizing to unknown generators.} \Qian{
Removing the patch-aggregation preprocessing of input images and adopting full-image convolution instead leads to a slight decrease in average accuracy, reflecting the effectiveness of non-downsampled patch-aggregation processing on raw pixel images.
} \Huang{Removing the high-level branch reduces the visual encoder to a CNN-based architecture, which is insufficient for few-shot detection and leads to a sharp drop in accuracy (average ACC 58.73\%), confirming the poor generalization of purely CNN-based encoders.} Removing the semantic-level feature leads to a drop in accuracy, most notably in the detection of MidJourney and WildRF. The captions provide multi-modal textual features beyond visual cues, offering greater benefits on certain datasets. \Huang{When LoRA is removed, the detection accuracy drops by about 3\%.} In contrast, fine-tuning with LoRA allows the LHSDet to maintain only $0.23\%$ of trainable parameters. LoRA fine-tuning significantly reduces the number of trainable parameters, decreases model size, and enhances detection accuracy.  In summary, each component of our proposed method contributes effectively. Both the low-level and high-level visual branch are indispensable, jointly capturing visual features. 
\Qian{In addition to the component removal studies, we further examine several design choices for the visual and textual backbones, as shown in Table~\ref{tab_backbone}.
Regarding the visual encoder, SigLIP2 consistently outperforms both CLIP and DINO2 across all datasets 
by an average margin of 9.1 and 16.8 points over CLIP and DINO2, respectively, with the largest gains on unseen generators such as MJv6 and WildRF.
Regarding the textual encoder, replacing the Vicuna-7B LLM (used in AntifakePrompt~\cite{antifake}) with Phi3-mini while keeping the same SigLIP2 backbone yields comparable detection trends but reduces the model parameter count by approximately half.
}
And we compare the fusion mechanisms for integrating the two visual branches.
Within the proposed architecture, the low-level and high-level visual branch are integrated via cross-attention. To validate this design, we incorporate a gated fusion mechanism as an alternative. When the two visual branches are integrated via gated fusion, the accuracy decreases by about $1\%$, with a particularly noticeable drop on MidJourney. 

\section{Conclusion}
\label{sec:con}
In this paper, we introduce LHSDet, \Qian{ 
a detector} for AI-generated images. 
It adopts a visual question answering (VQA) framework, 
leveraging the complementary strengths of the low-level and high-level visual information as well as the textual semantic information. The proposed framework processes input images through three complementary branches: a low-level visual branch performs non-overlapping patch aggregation to extract local texture features including SRM noise and neighboring pixel relationships, a high-level visual branch utilizes SigLIP2 to capture global perception concepts, and a semantic-level textual branch generates captions via BLIP-2. The two visual branches are further integrated through cross-attention mechanisms. Then the fused \Huang{multi-modal} representation is processed by the LLM yielding an answer. Extensive experimental results demonstrate that LHSDet achieves state-of-the-art detection accuracy of \Huang{96.41\% on average} across diverse datasets covering diffusion models, autoregressive models, and the WildRF benchmark. Furthermore, LHSDet exhibits strong robustness under different image post-processing operations.

\Huang{Despite its strong performance, LHSDet has limitations. First, the reliance on BLIP-2 captions introduces additional inference cost.
Second, the accuracy on MidJourney v6 (85.33\%) is relatively lower than on other test sets, suggesting that highly aestheticized imagery remains challenging.}
In future work, the triple-branch multi-modal feature extraction capability of LHSDet could be extended to other VQA tasks, thereby expanding its applicability beyond AI-generated image detection. Additionally, we aim to further enhance the model’s generalization ability to newly emerging generative models, addressing the long-standing challenge of detector adaptability in dynamic AI image generation landscapes.

\bibliographystyle{IEEEtran}
\bibliography{references}

\vfill

\end{document}